\pdfoutput=1

\documentclass[11pt]{article}

\usepackage[]{acl}

\usepackage{times}
\usepackage{latexsym}

\usepackage{pbox}
\usepackage{makecell}
\usepackage{graphicx}
\usepackage{subcaption}
\usepackage{setspace}
\usepackage{enumitem}
\usepackage{booktabs, cellspace, multirow}
\usepackage{amsmath}
\usepackage{amssymb}
\usepackage{soul}
\usepackage{xspace}
\usepackage{caption}
\usepackage{tabularx}
\usepackage{makecell}

\newcommand\Tstrut{\rule{0pt}{2.2ex}}         

\usepackage[T1]{fontenc}

\usepackage[utf8]{inputenc}

\usepackage{microtype}

\usepackage[T1]{fontenc}

\usepackage[utf8]{inputenc}

\usepackage{arydshln}
\usepackage{xurl}

\usepackage{microtype}

\newcommand{\metricname}{\textsc{Target-Coherence}\xspace}

\newcommand{\coda}{\textsc{CODA}}
\newcommand{\codanocskb}{\textsc{CODA-OnlyDA}}
\newcommand{\codanoda}{\textsc{CODA-NoDA}}
\newcommand{\codakw}{\textsc{CODA-NoEdge}}
\newcommand{\codaupper}{\textsc{CODA-Upper}}
\newcommand{\codaps}{\textsc{CODA-KBPath}}
\newcommand{\codanoalign}{\textsc{CODA-NoAlign}}

\newcommand{\dataurl}{www.github.com/prakharguptaz/target-guided-dialogue-coda}

\title{Target-Guided Dialogue Response Generation Using Commonsense and Data Augmentation}

\author{Prakhar Gupta$^\clubsuit$ \quad Harsh Jhamtani$^\clubsuit$ \quad Jeffrey P. Bigham$^{\clubsuit,\heartsuit}$  \\
$^\clubsuit$Language Technologies Institute, Carnegie Mellon University \\
$^\heartsuit$Human-Computer Interaction Institute, Carnegie Mellon University \\
\texttt{\small prakharg@cs.cmu.edu, jharsh@alumni.cmu.edu, jbigham@cs.cmu.edu}}

\begin{document}
\maketitle
\begin{abstract}

Target-guided response generation enables dialogue systems to smoothly transition a conversation from a dialogue context toward a target sentence. Such control is useful for designing dialogue systems that direct a conversation toward specific goals, such as creating non-obtrusive recommendations or introducing new topics in the conversation.
In this paper, we introduce a new technique for target-guided response generation, which first finds a bridging path of commonsense knowledge concepts between the source and the target, and then uses the identified bridging path to generate transition responses.
Additionally, we propose techniques to re-purpose existing dialogue datasets for target-guided generation.
Experiments reveal that the proposed techniques outperform various baselines on this task.
Finally, we observe that the existing automated metrics for this task correlate poorly with human judgement ratings.
We propose a novel evaluation metric that we demonstrate is more reliable for target-guided response evaluation.
Our work generally enables dialogue system designers to exercise more control over the conversations that their systems produce.\footnote{Code available at \url{\dataurl}}
\end{abstract}

\section{Introduction}
\label{sec:intro}

\begin{figure}[tb]
    \centering
    \vspace{-.2pc}
    \includegraphics[width=0.40\textwidth]{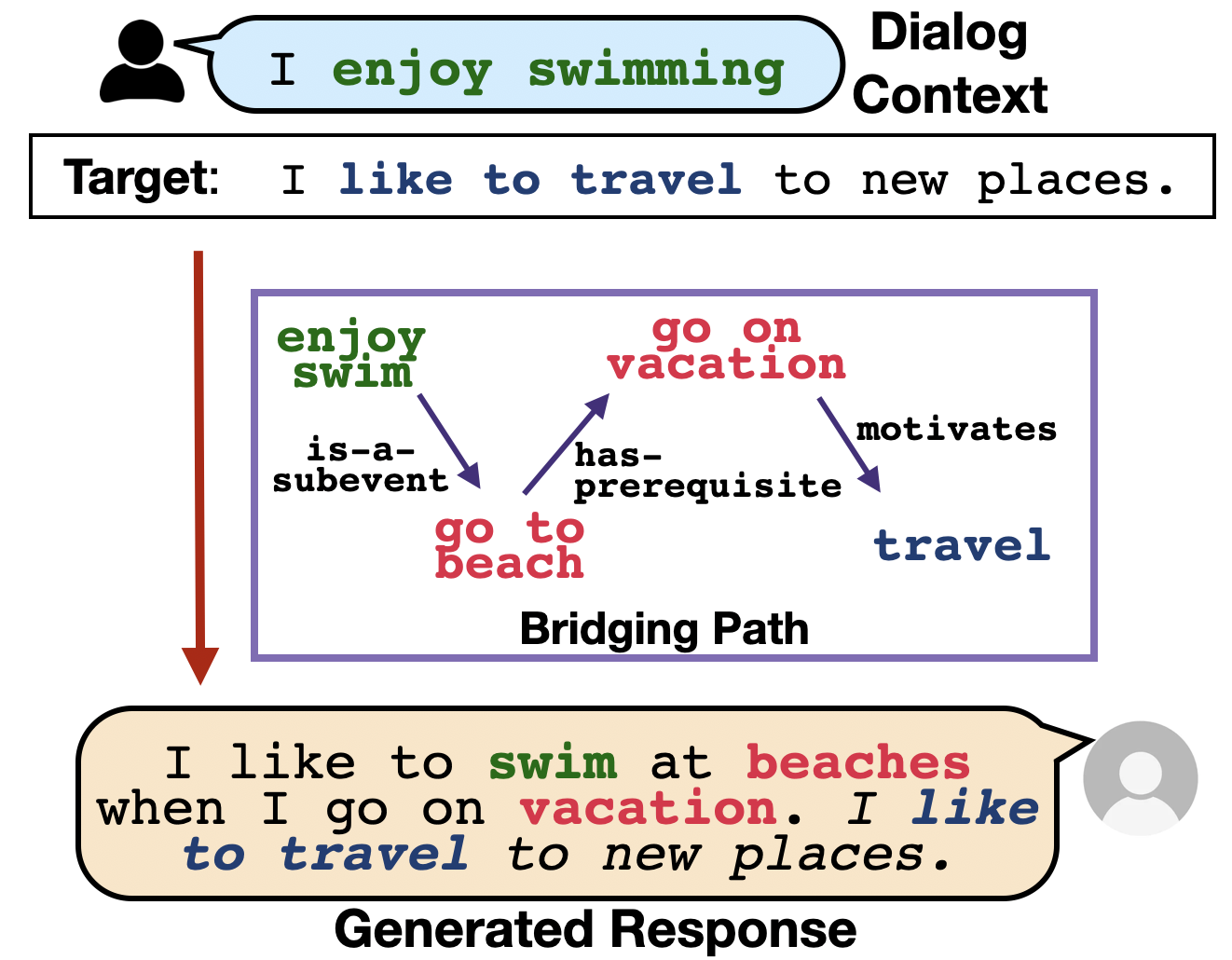}
    \vspace{-.5pc}
    \caption{
    \small 
    Given a dialogue context and a target sentence, our goal is to generate a dialogue response that smoothly transitions the conversation from context towards the target. Our proposed approach involves identifying a bridging path of entities to link the context and the target. }
    \label{fig:pullfig}
\end{figure}

Open-domain conversational systems have made significant progress in generating good quality responses 
driven by strong pre-trained language models \cite{radford2019language, devlin-etal-2019-bert} and large-scale corpora available for training such models.
However, instead of passively responding to a user, dialogue systems can take on a more proactive role to make recommendations, help users discover new services, or introduce interesting new topics to users to improve user experience. 
Furthermore, a proactive or target-guided system can guide the conversation towards safer conversational topics in case a conversation goes awry or a user becomes abusive towards the system, and direct the users towards topic areas that the system knows how to talk about.
Prior work has used mechanisms such as emotion labels~\cite{zhong2019affect}, persona~\cite{Song2019ExploitingPI}, and politeness \cite{niu2018polite} to control conversations. However, such approaches require labeled training data for a set of pre-determined labels, making it harder to incorporate new goals into a system.
In this work, we study the problem of proactive response generation based on a target sentence.
For example in Figure \ref{fig:pullfig}, given the context `I enjoy swimming', the system guides the conversation towards the target `I like to travel to new places' by mentioning `I like to swim at beaches when I go on vacation'. Using target sentences for proactive control is an intuitive and flexible control mechanism for dialogue developers, free of domain-specific handcrafting and annotations.

Existing publicly available dialogue corpora generally consists of free-flow conversations where the speakers move the conversation forward based on the dialogue history alone, with no particular agenda. We build upon the recently released \textit{Otters} dataset~\cite{DBLP:conf/acl/SevegnaniHKR20} with one-turn topic transitions for mixed-initiative in open-domain conversations. 
Given a source sentence from a speaker, the task is to generate a topic transition sentence with “bridging” strategies to a target sentence from another speaker. The task is challenging on several fronts. 
First, the system needs to balance the trade-off between coherence with the context while smoothly transitioning towards the target. 
Second, the Otters training dataset is relatively small (less than 2000 training instances), 
making it a low-resource setting.
Finally, we show that standard word-overlap metrics are insufficient for this task.

In this work, we propose methods to leverage commonsense knowledge from ConceptNet~\cite{speersConceptnet} to improve the quality of transition responses. 
Our technique decomposes the response generation process into first generating explicit commonsense paths between the source and target concepts, followed by conditioning on the generated paths for the response generation. This is intended to mimic how humans might bridge concepts for creating transitions in conversations using commonsense knowledge.
This technique offers two benefits: 1) Leveraging external ConceptNet knowledge solves the data scarcity issue and improves the model's capability to generate logical transitions; 
2) Since the transition response is grounded on commonsense knowledge paths, the explicit paths used by the model can provide explanations for the concepts used by the model, as well as provide control over the generation process.
Furthermore, we propose a data augmentation mechanism to help with the data scarcity issue by re-purposing training data from DailyDialog, an open-domain dialogue dataset. 
Both these approaches are complementary and outperform existing baselines in response quality and transition smoothness.
We demonstrate how the proposed approach of using explicit bridging paths enables improved quality of transitions through qualitative and human studies.

Automated evaluation is a challenging aspect of dialogue response generation tasks \cite{ZhaoZE17}. 
We show that the existing word-overlap metrics such as BLEU can be easily fooled to assign high scores to poor responses just based on high n-gram overlap with reference responses. We propose a metric \metricname which is trained using hard adversarial negative instances and achieves a high correlation with human judgement ratings of system outputs. As part of this work, we collect and release a dataset of human ratings of various system outputs for this task. 

We discuss the broader impact and potential uses of the proposed system, its limitations and potential ethical issues related to this task in Section~\ref{sec:ethics}.

\section{Related Work}

\noindent \textbf{Target Guided Dialogue Response Generation:}
\citet{DBLP:conf/acl/SevegnaniHKR20} is perhaps the closest to our work described in this paper. They work on the task of generating a new utterance which can achieve a smooth transition between the previous turn's topic and the given target topic. 
Past work in controllable text generation has explored steering neural text generation model outputs to contain a specific keyword \cite{DBLP:journals/corr/abs-1909-05858}, a knowledge graph~\cite{wu-etal-2019-proactive}, or a topic \cite{DBLP:journals/ipm/LingCHLCC21}.
Steering dialogue towards a given keyword has also been explored in past work \cite{DBLP:conf/acl/TangZXLXH19,DBLP:conf/aaai/QinYTL20, Zhong_Liu_Wang_Miao_2021}, albeit as a retrieval task. 
In contrast, our goal is to generate a next utterance in a dialogue setup which can steer a conversation towards target sentence in a smooth fashion rather than generating a response for a given keyword or topic.
Our work is also related to prior work on text infilling \cite{DBLP:conf/acl/DonahueLL20,DBLP:conf/emnlp/QinSWBHBBC20}, though compared to them we work in a dialogue setup and utilize commonsense knowledge to perform the infilling.

\noindent \textbf{Commonsense for Dialogue Generation:}
Commonsense knowledge resources \cite{speer2017conceptnet,malaviya2020commonsense} have been used in dialogue response generation for tasks such as persona-grounded dialogue  \cite{DBLP:conf/emnlp/MajumderJBM20} and open-domain dialogue generation \cite{DBLP:conf/aaai/GhazvininejadBC18, hedayatnia-etal-2020-policy, zhou-etal-2021-think}.
\citet{zhou-etal-2021-commonsense} created a dataset
focusing on social commonsense inferences in dialogue and \citet{arabshahi2020conversational}  designed a
theorem prover for if-then-because reasoning. A concurrent work~\cite{zhou2021think} proposed to train a model to explicitly generate implicit knowledge and use this knowledge to generate a response. Compared to their work, we focus on target-guided response generation, suggest mechanism for knowledge alignment with the transition response during training, and focus on multi-hop knowledge paths. 
More broadly, commonsense knowledge has been used in text generation tasks such as story and essay generation \cite{DBLP:conf/aaai/GuanWH19,DBLP:conf/acl/YangLLLS19}.
%

\noindent \textbf{Automated Metrics for Evaluating Dialogue Quality:}
Automated metrics such as BLEU \cite{papineni2002bleu}, METEOR \cite{banerjee2005meteor}, and BertScore \cite{DBLP:conf/iclr/ZhangKWWA20} are widely used to evaluate quality of machine-generated text. 
However, such metrics often correlate poorly with human judgement ratings of generated text quality \cite{sai2020survey}.
Past work has explored trained model-based metrics such as ADEM \cite{lowe2017towards} and RUBER \cite{tao2017ruber}. However, training such model-based metrics often relies on tagged training data. \citet{DBLP:conf/acl/GuptaTB21} propose ways to mitigate the need for such labelled data by automatically synthesizing negative examples. Our proposed metric is along similar lines, though we utilize different techniques for synthetic negative example generation.

\begin{figure*}[tb]
    \centering
    \vspace{-.2pc}
    \includegraphics[width=0.76\textwidth]
    {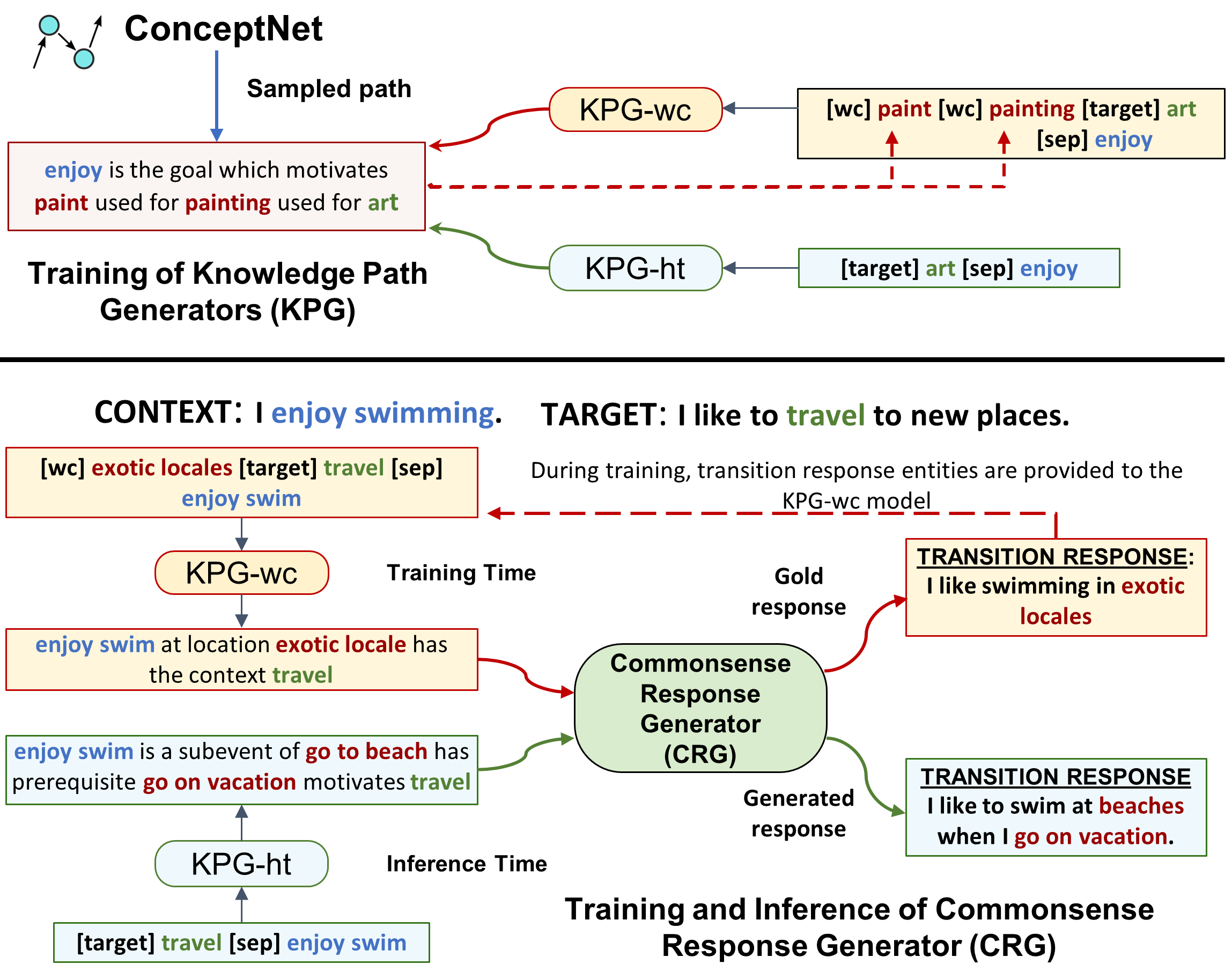}
        \caption{ \small Model illustrations for KPGs - Knowledge Path Generators (top) and CRG - Commonsense Response Generator (bottom). Base architecture for all models is GPT-2. Given a path sampled from ConceptNet, KPG-wc learns to predict the path given the head, tail and intermediate entities of the path while KPG-ht learns to predict the path given only the head and tail entities.  
        For the CRG model, during training, a head entity from the context, a tail entity from the target and intermediate entities from the gold transition response are fed into KPG-wc and its output path is used as input to the CRG model. During inference, a head entity from the context and a tail entity from the target are fed into the KPG-ht model. KPG-ht then generates a path with new concepts such as ``go on vacation''. CRG model conditions on this path for transition response generation.   }
    \label{fig:interface1}
     \vspace{-1em}
\end{figure*}

\section{Task Overview}
\label{sec:background}
We first formalize the task of target-guided response generation. 
Given a conversation context $c$ between two speakers A and B, and a target utterance $t$ for speaker B, the task is to generate a transition sentence $s$ which serves as a smooth link between the context and the target. The target is a phrase or a sentence.
\textit{Otters} dataset~\cite{DBLP:conf/acl/SevegnaniHKR20} consists of a simplified setting of one-turn topic transitions, where the conversation history consists of a single utterance $u_a$ from speaker A, and a target utterance $u_b$ for speaker B, and the task is to generate a transition utterance $s$ for speaker B to serve as a smooth link between $u_a$ and $u_b$. 
The task is challenging since a system needs to devise a strategy that balances the competitive objectives of generating a response which is coherent to the context, while smoothly driving the conversation towards the target.


In this work, we propose two approaches for transition response generation task: 1) Commonsense-guided response generation (\autoref{sec:commonsense}), and 2) Data augmentation to tackle data sparsity (\autoref{sec:augmentation}).
We refer to the proposed method as \textbf{\coda{} (Commonsense Path and Data Augmentation)}.
We also propose a novel metric \textbf{\metricname{}} to automatically evaluate smoothness of response transitions (\autoref{sec:tcmetric}).

\section{Commonsense-Guided Response Generation}
\label{sec:commonsense}
We frame the target-guided response generation task as follows. 
Given a conversation context $c$ and a target $t$, a conditional language model learns to predict the transition response $s$.
Target-guided generation can potentially benefit by incorporating commonsense reasoning by 
 identifying rich connections between a pair of entities which enable us to generate logical transition responses connecting the two.
Pre-trained language models are 
known to suffer in cases where commonsense knowledge is required during generation~\cite{ijcai2018-643, Guan_Wang_Huang_2019}, especially in tasks where there is not enough data available for learning commonsense patterns from the text, which is true for our case.
In contrast, Commonsense Knowledge Graphs 
like ConceptNet~\cite{speersConceptnet} provide structured
knowledge about entities, which enables higher-level reasoning about concepts. 
%

In this work we use commonsense knowledge from ConceptNet for planning a 
transition response. ConceptNet is a large-scale semantic graph that has concepts as nodes and has commonsense relationships between them, such as `IsA' and `AtLocation'.
However, ConceptNet suffers from severe sparsity issues~\cite{malaviya2020commonsense, bosselut-etal-2019-comet}. Therefore, it is not always possible to find the concepts and relationships between context and target concepts. 
To address the sparsity issue, we develop Knowledge Path Generator (\textbf{KPG}), 
a language model trained on paths sampled from ConceptNet.
The model takes a pair of entities or concepts as input 
and generates a multi-hop path connecting the two. 
Since the knowledge paths are sampled from a generative model rather than retrieved from a fixed knowledge base, we are no longer limited by the entities and paths present in the ConceptNet knowledge base.

To generate commonsense based responses, we train a Commonsense Response Generator (\textbf{CRG}) model to generate the transition response conditioned on the paths generated by the KPG model (\textbf{Figure} \ref{fig:interface1}). Conditioning the response generation on commonsense paths improves the reasoning capabilities of the CRG model and provides the added benefits of interpretability and control over the generation process.

\subsection{Commonsense path generator}
\label{sec:pathgenerator}
The KPG models attempts to connect a concept or entity phrase from the context to a concept from the target by creating knowledge paths between them.

\noindent
\textbf{Path Sampling:} 
To create training data for the KPG models, we sample paths between entity phrases from ConceptNet using random walks. This step builds upon past work of \citet{wang-etal-2020-connecting}.
Given nodes $N$ and edges $E$ from ConceptNet, we perform random walks on the graph to sample a set of paths $P$ of the form $p=\{n_0,e_0,n_1,e_1,...,e_{k-1},n_k\}\in P$. Here, a path $p$ connects a head entity phrase $n_0$ with the tail entity phrase $n_k$ via intermediate entities and edges (or relations) $n_i, e_i$. 
To sample paths, the random walk begins with a random entity node $n_0$ and samples a path of random length $k \in \{1,2,...,K\}$, where we have set $K=6$ in this work.
To sample paths that are useful for our task, we prevent sampling certain edges types such as \textit{Synonym} (Appendix~\ref{sec:edges}). 

\noindent
\textbf{KPG-head-tails (KPG-ht):}
KPG-ht is a GPT-2~\cite{radford2019language} based model which is trained to predict a knowledge path
$p$ which links a head entity $n_h$ to a tail entity $n_t$. 
For a sample path $p=\{n_h,e_0,n_1,e_1,...,e_{k-1},n_t\}$ from ConceptNet, 
the path is formatted into the following sequence ``[target] $n_t$ [sep] $n_h$ $e_0$ $n_1$ $e_1,…,e_{k-1}$ $n_t$''. 
KPG-ht is only used during CRG inference where the head entity is extracted from the context and tail entity from the target (Figure ~\ref{fig:interface1}). 

\noindent \textbf{KPG-will-contain (KPG-wc):}
A large number of possible paths can exist for a given head-tail entity pair.
Training the CRG model by conditioning on paths which are irrelevant to the gold transition response might discourage the CRG model from conditioning on the provided commonsense path.
Since we do not have gold paths for a response, we instead train a model KPG-wc to generate paths which are more aligned to the gold response by enforcing the generated path to contain entities from the gold response.
%
{KPG-wc} is trained to predict a path which contains a pre-specified entity set $E_p = \{k_1,...,k_n\}$ in the generated path by formatting paths sampled from ConceptNet as the following sequence: ``[wc] $k1$ [wc] $k2$…[target] $n_t$ [sep] $n_h$ $e_0$ $n_1$ $e_1,…,e_{k-1}$ $n_t$'' (Figure ~\ref{fig:interface1}).
The entity set $E_p$ is a randomly permuted sequence of entities  $n_1, n_2,…, n_{k-1}$ from the sampled path. Here ``wc'' symbolizes ``will contain''. 
Training with this sequence indicates to the model that the path generated between $n_h$ and $n_t$ should contain the entities from the set $E_p$ in a sensible order. Specifying the special token ``[target]'' followed by the tail entity $n_t$ informs the model about the last entity it should output when generating a path. We discuss how the set $E_p$ is constructed for  CRG model training in the next section.

In practice, we train a single common GPT-2 based model for KPG-wc and KPT-ht. The model at test time is able to generate knowledge paths for either case, whether in-path entities from $E_p$ are present (KPG-wc) in the input or not (KPG-ht).

\subsection{Response generator}
\label{sec:responsegenerator}
The Commonsense response generator conditions on the commonsense  paths generated from the KPG models to generate the transition responses. 

\vspace{1mm}
\noindent
\textbf{Entity extraction}.
We extract a set of entities $E_h, E_t$ and $E_r$ from the context, target and gold transition response respectively using NLTK.
We designed simple grammar rules (details in Appendix \ref{sec:codadetails}) to convert phrases to concise forms that match the nodes present in ConceptNet, {\em e.g.}, ``watching the star'' is converted to ``watch stars''. 

\vspace{1mm}
\noindent
\textbf{Sampling and filtering paths:}
In this step, for every pair of head and tail entity from $E_h$ and $E_t$, we sample multiple paths from the KGP models using topk sampling and chose one or more of these paths for training and inference. 
\textit{For training} the CRG models with the commonsense paths, we need to curate paths that are relevant to and aligned with the gold response so that they are not ignored by the CRG model during inference. We achieve this by first sampling paths which are relevant to the gold response, and then apply filtering mechanisms to curate the final set of paths. For training data path sampling, we use the \textit{KPG-wc} model (Figure~\ref{fig:interface1}). The input to the model is a head and tail entity pair $n_h$ and $n_t$, and the entity set $E_p$ that consists of the set of entities $E_r$ from the gold transition response. The model then generates a set of paths that contain the head and tail entities as well as the gold response keywords. Thus, the sampled path is inherently relevant to the gold response due to the conditioning on gold keyword entities. \textit{During inference}, the set $E_r$ is not available, so we leverage the  \textit{KPG-ht} model that takes just the head and tail entity pair $n_h$ and $n_t$ as input to generate a commonsense path.

Assuming the context and target consists of $m$ and $n$ entities each, and we generate $q$ number of paths per pair, we get a total of $m\times n \times q$ number of paths for each data instance.
Since $m\times n \times q$ can be a large number, we use simple methods to sub-select entity pairs and paths. 
\textbf{(1)} Sub-selecting Entity Pairs: We score an entity pair 
by calculating the inverse document frequencies (computed using Gutenberg English corpus) of the entity tokens and summing up the maximum value found for a token in each entity in the pair. For training phase, we keep the top D pairs of entities, and for testing phase we keep only the highest-scoring pair.
\textbf{(2)} Sub-selecting paths: We apply the following strategies to prune the set of paths for each entity pair: 1) \textit{Perplexity} - We filter out all the paths whose perplexity values (from the KGP models) are more than double the average perplexity values of all paths between an entity pair. 2) We remove all the paths which have repetition of entities since repetition often leads to degeneration during decoding. 3) For paths in training data, we filter out paths which contain entities not present in the gold response. 
The final set of paths $P$ are converted into natural language by converting the relation and inverse relations into textual format. For example, ``art gallery UsedFor for art'' is converted to ``art gallery is used for art''.

\vspace{1mm}
\noindent
\textbf{Training and inference in CRG model}. The CRG model (GPT-2 based) is trained as a conditional model with following input sequence: ``\textit{knowledge path} [target] \textit{target sentence} [context] \textit{context sentence} [response] \textit{transition response}'' for each \textit{knowledge path} from set $P$. We train CRG model by minimizing the log-likelihood loss of the transition response.
For inference, we create the set of paths $P$ by entity extraction, path sampling and filtering and choose a random path $p$ from final set $P$. The model generates transition response conditioned on the sequence $c, t$, and $p$.

\section{Data Augmentation}
\label{sec:augmentation}
The task of target-guided response generation is still a relatively unexplored task, and Otters~\cite{DBLP:conf/acl/SevegnaniHKR20} is the only suitable dataset for this task to the best of our knowledge. 
However, Otters is small and consists of only a few hundred context-target pairs.
This makes learning transition concepts and strategies challenging in this low-resource setup. On the other hand, there are many publicly available dialogue datasets for training response generation models.
Such datasets contain free-flow conversations, where although the speakers generate context coherent responses, they do not condition their responses on any target. We propose a technique to leverage and re-purpose such datasets for the task of target-guided response generation. We pick the DailyDialog~\cite{li-etal-2017-dailydialog} dataset for experimentation and convert its conversations to target-guided conversations in two steps: 1) Target creation, and 2) Data filtering.

\begin{table}[t]
\centering
\small
\begin{tabular}{lp{0.32\textwidth}}
\toprule
Context       & the restaurant looks authentic european.                   \\
Response      & the chef trained in florence. the pasta tastes nice here.  \\
SRL Output    & predicate = tastes, arguments= the pasta; nice here        \\
Target clause & the pasta tastes nice here. \\
\bottomrule
\end{tabular}
 \captionof{figure}{
    \small An example to demonstrate how a conversation in DailyDialog can be re-purposed for the task of target-guided response generation. 
    }
    \label{fig:aug}
\end{table}

For \textit{target creation}, we run Semantic Role Labelling (SRL) to predict predicate and arguments in a response. For each predicate identified, we create a clause by putting together the predicate and arguments in a textual sequence. Finally, we only use the clause occurring towards the end of the response as a target. 
An example for target creation is shown in Figure \ref{fig:aug} (More details about clause identification are in Appendix~\ref{sec:clause}).

The target creation step does not guarantee that a candidate response transitions smoothly towards the target clause. 
In the \textit{data filtering} step, we introduce a \metricname metric to score a transition response in terms of its coherence to the context and smoothness towards the target. The metric is described in more detail in section~\ref{sec:tcmetric}. The metric assigns a score between 0-1 for a transition response and we remove instances with a score less than a threshold $k$ (set to 0.7) from consideration. The remaining instances are used for pretraining response generation models which are finally fine-tuned on the Otters dataset.

\section{Target-Coherence Metric} 
\label{sec:tcmetric}
Evaluating target-guided responses is a challenging task as a good transition response needs to be both - coherent to the context and smoothly transition towards the target. Furthermore, since the task is open-domain and open-ended, there are many possible correct responses which may not match with a reference response~\cite{elikyilmaz2020EvaluationOT}. To tackle  these challenges, we propose an automatic metric for this task that does not use human references. The proposed metric \textbf{\metricname} is based on a classification model trained to classify a transition response as either \textit{positive}, that is, it is coherent to the context and smoothly transitions towards the target, or negative, that is, the response is either not coherent to the context or does not transition towards the target.


\begin{table}[t]
\centering
\addtolength{\tabcolsep}{-4pt}
\small
  \setlength\extrarowheight{-2pt}
\renewcommand{\arraystretch}{0.7}
\begin{tabular}{l|l|l} 
\hline
\multirow{3}{*}{\begin{tabular}[c]{@{}l@{}}\textbf{POSITIVE}\\Gold c,r,t\end{tabular}} & CONTEXT c & \begin{tabular}[c]{@{}l@{}}the restaurant looks authentic\\~european.\end{tabular} \\ 
\cdashline{2-3}
 \Tstrut
 & RESPONSE r & \begin{tabular}[c]{@{}l@{}} the chef trained in florence.\end{tabular} \\ 
\cdashline{2-3}
 \Tstrut
 & TARGET t &the pasta tastes nice here. \\ 
\hline
\begin{tabular}[c]{@{}l@{}} \Tstrut \textbf{NEGATIVE}\\Random t' \\with gold r,c\end{tabular}         & TARGET t'  & i love to drive my  car.                                  \\ 
\hline
\begin{tabular}[c]{@{}l@{}}\Tstrut \textbf{NEGATIVE}\\Random c' \\with gold r,t\end{tabular}         & CONTEXT c' & i enjoy computers and
  phones.                            \\
\hline
\begin{tabular}[c]{@{}l@{}}\Tstrut \textbf{NEGATIVE}\\Random r' \\with gold c,t\end{tabular}         & RESPONSE r' & there is no parking here.                            \\
\hline
\end{tabular}
 \captionof{figure}{
    \small We train a reference-less model-based metric \metricname{} to score the smoothness of a generate response wrt to dialogue context and target sentence. To train the metric, we synthesize hard negative examples using an ensemble of techniques as shown in this figure. 
    }
    \label{fig:tcsynth}
    \vspace{-.2pc}
\end{table}

We use the gold transition response from the training dataset to create positive instances for training. For a positive instance with context $c$, target $t$ and response $r$, we create negative instances using the following mechanisms: 
1) We hold two out of (c,t,r) constant while randomly sample the third one. For example, sample a random context $c'$, which makes $r$ incoherent to the $c'$. An example is shown in Figure~\ref{fig:tcsynth}.
2) We use a GPT-2 model trained on Otters dataset to generate a response $r'$  coherent to $c$ but conditioned on a random target $t'$. 
3) For a target $t$, we chose a response $r'$ from the Otters training set which has $t$ as the target but context $c' \neq c$. 
We sample a maximum of 2 negative instance per mechanism and balance the count of positive and negative instances by  repeating positive instances. 
We fine-tune a pre-trained BERT-base~\cite{devlin-etal-2019-bert} model on these instances with binary cross entropy loss. 
\section{Experiments}
\label{sec:experiments}

\begin{table}[t]
\centering
\small
\begin{tabular}{lccc}
\toprule
Dataset     & Train       & Dev         & Test        \\
\hline
 \Tstrut
Otters-id   & 1,929 (693) & 1,160 (404) & 1,158 (303) \\
Otters-ood  & 2,034 (677) & 1,152 (372) & 1,130 (372) \\
DailyDialog & 11,118      & 1,000        & 1,000    \\
\bottomrule
\end{tabular}
 \caption{Overview of the datasets.}
    \vspace{-3mm}
    \label{tab:datasets}
\end{table}

\subsection{Datasets}
We use two datasets in our experiments. 1) Otters~\cite{DBLP:conf/acl/SevegnaniHKR20} contains instances with context-target-transition response triplets. It consists of two sets of splits. The Out-Of-Domain (\textsc{OOD}) split ensures that none of the context-target pairs in the test set are present in the train set.
In the In-Domain (\textsc{ID}) split, one of either the context or the target in each pair in the test-set is allowed to appear in the
train-set. DailyDialog dataset consists of casual conversations between two speakers. In Table~\ref{tab:datasets} we present the number of dialogues in DailyDialog dataset and number of responses in otters, along with number of unique context-target pairs in brackets. Otters dataset consists of multiple responses per context-target pair. Some transition responses in Otters dataset are noisy - they contain sentences and phrases from the target sentences. We remove such data from the test sets (with word overlap $>$ 0.75), leaving 1019 data points in the Otters-id test set and 988 data points in the Otters-ood test set.

\subsection{Baselines for generation}


We report results for a number of {baselines}. 
We provide complete implementation details of \textsc{CODA} and all baselines in Appendix~\ref{sec:addtional} and \ref{sec:baselinedetails}.
\begin{itemize}[noitemsep,nolistsep, leftmargin=*]
  \item \textbf{GPT-2:} \cite{radford2019language} A pretrained GPT--small language model fine-tuned on Otters data. Conditions on the context and target sentences to generate the transition response.
    \item  \textbf{GPT2-Fudge} \citet{DBLP:conf/naacl/YangK21} 
  uses a discriminator trained to distinguish good response continuations from the poor ones and guides the GPT-2 based decoder towards responses that are coherent to both the source and target sentences.
  \item  \textbf{Multigen} \cite{DBLP:conf/emnlp/JiKHWZH20} 
  combines the vocabulary distribution generated by underlying GPT-2 model with a concept distribution from a commonsense knowledge base (ConceptNet).
    \item \textbf{Concept-Predict}  leverages  a  concept  prediction  strategy  from \citet{DBLP:conf/aaai/QinYTL20}.  The concept is predicted based on closeness to the target. 
  \item \textbf{CS-Pretrain} model is pretrained with commonsense paths used for training the KPG models and is based on the commonsense story generation model from \citet{Guan2020AKP}.
\end{itemize}

\noindent
\textbf{Ablation experiments}: We report results for following {\textsc{CODA} {variants}}:
\begin{itemize}[noitemsep,nolistsep, leftmargin=*]
  \item \textbf{\codanocskb{}}: \coda{} variant that uses DailyDialog augmentation and does not use commonsense paths from KPG models in the CRG model. 
  \item \textbf{\codanoda{}}: \coda{} trained without additional data from DailyDialog.
  \item  \textbf{\codakw{}} \coda{} variant that uses only entities and no edges from the path.
  \item \textbf{\codanoalign{}}: variant that relies on only KPG-ht for training and inference. Does not select paths based on alignment with responses.
   \item \textbf{\codaps{}}: variant that retrieves paths directly from ConceptNet  using the algorithm proposed in \citet{lin-etal-2019-kagnet}.
  \item  \textbf{CODA-Upper} Upper bound for \coda{} which uses paths inferred from the gold responses using the KPG-wc keywords model during inference. 
\end{itemize}

\begin{table}[tb]
\centering
\addtolength{\tabcolsep}{-5pt}
\small
\begin{tabular}{l|ccc|c}
\toprule
\textbf{Metric} & \begin{tabular}[c]{@{}c@{}}\textbf{Target as} \\ \textbf{response}\end{tabular} & \begin{tabular}[c]{@{}c@{}}\textbf{Context as} \\ \textbf{response}\end{tabular} & \begin{tabular}[c]{@{}c@{}}\textbf{Reference}\\  \textbf{response}\end{tabular} & \begin{tabular}[c]{@{}c@{}}\textbf{Correlation} \\ \textbf{w ratings}\end{tabular} \\ \hline
\Tstrut
BLEU & 15.0 & 9.9 & 6.5 & -0.11 \\
METEOR & 14.0 & 12.6 & 13.2 & { 0.01} \\
ROUGE-L & 32.3 & 29.8 & 26.5 & -0.04 \\
BS-rec & 38.1 & 38.9 & 41.3 & { 0.05} \\
BS-F1 & 42.8 & 42.6 & 38.9 & -0.06 \\
\pbox{20cm}{\textsc{Target}-\\\textsc{Coherence}} & 10.7 & 4.0 & 77.4 & \underline{ 0.47} \\\bottomrule
\end{tabular}
 \caption{
 \small 
 We present the metric scores when using the target, context and one of the references as the response. All metrics except for  \metricname score the target and context higher than the reference. \metricname achieves high correlation with human ratings. Underlined values represent statistically significant result with p-value$<$0.05.}
    \vspace{-4mm}
    \label{tab:evalcorr}
\end{table}

\begin{table*}[]
\centering
\addtolength{\tabcolsep}{-1pt}
\small
\renewcommand{\arraystretch}{0.77}
\begin{tabular}{lccccc|ccccccc}
\toprule
 & \multicolumn{5}{c|}{In-Domain} & \multicolumn{5}{c}{Out-Of-Domain} \\
 \hline
 \Tstrut
 & BLEU & METEOR & ROUGE-L & BS-rec  & TC & BLEU & METEOR & ROUGE-L & BS-rec  & TC \\
 \hline
  \Tstrut
GPT-2 & 3.4 & 11.9 & 23.9 & 35.4 & 26.7 & 3.0 & 10.8 & 22.2 & 35.0 & 29.7 \\
GPT2-Fudge & 3.4 & 12.4 & 24.4 & 36.1 & 28.3 & 3.4 & 11.1 & 23.0 & 35.1 & 29.6 \\
Multigen & 6.2 & 12.5 & 28.1 & 40.0 & 27.8 & 4.9 & 11.6 & 26.0 & 36.7 & 30.8 \\
{Concept-predict} & 3.3 & 12.3 & 28.5 & 38.1 & 28.3 & 3.7 & 11.6 & 23.1 & 35.9 & 26.3 \\
CS-Pretrain & 2.8 & 11.1 & 23.2 & 35.2 & 21.5 & 2.8 & 10.2 & 21.2 & 33.0 & 22.0 \\
\textsc{CODA} & 5.0 & 12.6 & 25.9 & 38.0  & \textbf{36.7} & 4.6 & 11.5 & 24.3 & 35.5 & \textbf{37.9}\\
\hdashline
 \Tstrut
\codanocskb & 4.0 & 12.4 & 24.4 & 37.5  & 32.7 & 3.1 & 11.1 & 22.7 & 35.3  & 33.2 \\
\codanoda & 4.4 & 12.3 & 25.1 & 37.8 & 35.7 & 4.5 & 11.6 & 24.4 & 35.4  & 36.0 \\
\codakw & 4.2 & 12.0 & 25.0 & 37.4  & 33.7 & 4.0 & 11.8 & 24.2 & 35.4  & 35.9 \\

\codanoalign & 3.7& 12.4 &  25.5&  38.5&  32.1& 3.2 & 11.2 & 22.8 & 35.6  & 31.2 \\
\codaps &3.6 & 12.5 & 24.9 & 38.6 & 33.9 &
3.6 & 11.4 &24.1 & 35.9 & 33.0 \\
\codaupper & 8.3 & 18.1 & 32.6 & 44.4  & 47.9 & 7.5 & 17.9 & 30.7 & 42.7  & 45.4 \\
\hdashline
 \Tstrut
Human & 6.5 & 13.1 & 26.5 & 41.3  & 77.4 & 4.9 & 12.3 & 24.0 & 37.6 & 77.3
\\
\bottomrule
\end{tabular}
    \vspace{-2mm}
 \caption{\small 
 We present the results of automatic evaluation based on word-overlap and proposed \metricname{}. 
 \coda{} outperforms all the baselines for most of the metrics. We also present results for CODA's model ablations.}
    \vspace{-3mm}
    \label{tab:evalmetric}
\end{table*}

\subsection{Evaluation Metrics}
\label{sec:evalbias}
We report standard automated metrics such as 
BLEU \cite{papineni2002bleu}, ROUGE-L \cite{lin2004rouge}, METEOR \cite{banerjee2005meteor}, and BertScore ({BS-rec} and {BS-F1}) \cite{DBLP:conf/iclr/ZhangKWWA20}. Evaluation is carried out using multiple references from the test set.
Word-overlap metrics do not correlate well with human judgements~\cite{liu-etal-2016-evaluate}. Additionally, we observe that on this task, even a poor transition response can get a high score on reference-based metrics if it has high overlap with the context or the target. 
We carry out an experiment where we use the target, context and one of the references as the transition response. An ideal metric would score the reference response high, and give low scores to target and context used as a response.
In Table~\ref{tab:evalcorr}, reference-based metrics assign higher scores to target and context sentences used as responses compared to human-written responses. 
In contrast, \metricname{} assigns high scores to reference responses and low scores to target and context sentences. 

\noindent \textbf{Correlation of metrics with human judgements:}
We investigate how well do the metrics correlate with human ratings of system outputs. 
To perform this analysis, responses from \coda{}, baselines, as well as reference responses are judged by crowd-source annotators who rate the smoothness of a response given the dialogue context and the target on a scale of 0 to 1.
We collect a total of 440 ratings across Otters ID and OOD splits, and report Spearman rank correlation \cite{spearman1961proof} of the metrics and the ratings. Krippendorff's alpha for annotation is 0.42. 
Results, shown in last column of \autoref{tab:evalcorr}, depict that most standard automated metrics correlate poorly with human ratings, while the, proposed \metricname{} achieves a very high correlation score of $0.47$.

We present the Amazon Mechanical Turk interface for human ratings collection in Figure~\ref{fig:mturkratings} in the Appendix. The workers were first shown instructions about the task with definitions and examples for all the rating criteria. 
We paid crowd workers on Amazon’s
Mechanical Turk platform \$0.7 per annotation and gave bonuses to annotators with high annotation quality. Our estimated hourly pay was \$13, which is above the minimum US federal hourly wage. We set the worker qualification condition as 1000 HITS completed, 95\% or more approval rate and location as native English speaking countries. 
We release the human ratings and system outputs used for computing the metric correlations as part of this work.

\subsection{Results}
In this section we present the automatic and human evaluation results.
Automated metric  results are summarized in \autoref{tab:evalmetric}.
Although reference-based metrics are lexically biased (subsection \ref{sec:evalbias}), we still report their scores.
We observe that \coda{} outperforms all the baselines under in-domain (ID) as well as out-of-domain (OOD) setups of Otters data as per \metricname{} (TC) score.
For example, \coda{} gets a high TC score of $36.7$ (ID) and $37.9$ (OOD) while the TC scores of the closest baselines GPT2-Fudge, Multigen and Concept-predict are in the range of $28$-$31$, demonstrating that the proposed method leads to significant improvements in response quality. However, CODA is far from reaching human performance (TC 77.4).

\textbf{\coda{} Ablations:} We observe that: 
(1) Not using commonsense knowledge (\codanocskb{}) leads to large performance drops, highlighting that \coda{} effectively utilizes commonsense knowledge.
(2) Dropping data augmentation leads to a small drop in performance (\codanoda{}), hinting at relatively small (but still significant) benefit from pretraining the model using data augmentation. 
(3) Low performance of \codakw{} shows the importance of using edges in commonsense paths.
(4) Not aligning and selecting paths based on their relevance to responses during CRG training (\codanoalign{}) leads to a high drop in performance.
(5)  CODA  outperforms \codaps{} by  8\% (ID) and 14.5\% (OOD). This improved performance can be attributed to the generalizability of entities and paths generated from the KPG models.
(6) \codaupper{} achieves high scores, highlighting that further improvement in commonsense path generation component can significantly boost the output quality of \coda{}.

\begin{table}[t]
\centering
\small
\begin{tabular}{llccc}
\toprule
          \textbf{Criteria} & \textbf{Models}       & \textbf{Win}       & \textbf{Lose}     & \textbf{Tie}     \\
                 \toprule
Smooth & CODA vs GPT-2    & 37.5      & 31.6     & 31.0    \\
& CODA vs Multigen & 32.3      & 22.8     & 44.8    \\
                \hline
                 \Tstrut
Sensible & CODA vs GPT-2    & 22.0     & 21.3     & 56.7    \\
& CODA vs Multigen & 25.8      & 25.6     & 48.6   \\
\hline \Tstrut
Informative & CODA vs GPT-2    & 32.3     & 27.3     & 40.4    \\
& CODA vs Multigen & 35.5      & 27.8     & 36.7   \\
\bottomrule
\end{tabular}
 \caption{
 \small 
 Human evaluation through pairwise comparison between CODA and baselines. CODA is preferred in smoothness and informativeness criteria while being comparably sensible.}
    \vspace{-3mm}
    \label{tab:humaneval}
\end{table}

\textbf{Human Evaluation:}
We conduct human evaluations on Amazon Mechanical Turk to evaluate the quality of generated transition responses. 
Annotators are requested to evaluate the transition response on following criteria: 
(1) \textit{Smooth}: rate whether the response serves as a smooth transition between the dialogue context and target.
(2) \textit{Sensible}: whether the response makes sense in itself i.e. it is grammatical and logically coherent.
(3) \textit{Informative}: how much informative content a response carries.
Human annotators compare (or mark as a tie) responses from two models.
We collect two annotations for 100 randomly selected data points from the test outputs.
Results in \autoref{tab:humaneval} demonstrate that \coda{} outputs are preferred over the baselines on `Smooth' and `Informative' criteria. 

\subsection{Qualitative Analysis}
We present representative outputs from the models in Table~\ref{tab:examples}. For CODA, we show the path used in response generation. We notice that GPT-2 and Multigen often tend to either generate simple outputs ({\em e.g.} `I hate my food' in the last example) or simply repeat or address either the target or the context ({\em e.g.} `My pet is the gecko', `Seattle is my favorite city to go.') which leads to high BLUE and METEOR scores, but low TC scores. CODA avoids these pitfalls as it is conditioned on  generated commonsense paths based on both the context and target entities leading to more informative and sensible outputs. 
However, CODA is susceptible to two issues: 1) Using poor keywords for path generation, and 2) Generation of irrelevant
paths ({\em e.g.} `server is a person not desires greasy food' in the last example).


\begin{table}[t]
\centering
\small
\begin{tabular}{l}
\toprule
\begin{tabular}[c]{@{}l@{}}\textit{Context}: i like the sand on my feet\\ \textit{Target}: my puppy is called georgie.\end{tabular}                                                                                                                                                                                    \\
\hdashline
\begin{tabular}[c]{@{}l@{}}
 \Tstrut
\textit{GPT-2}: My mom likes the water.\\ \textit{Multigen}: My pet is the gecko.\\ \textit{CODA}: My dog walks along the beach with sand.\\ \textit{CODA-Path}: sand is at location beach belongs to walk \\    is      desired by puppy\end{tabular}                                                             \\


\toprule
\begin{tabular}[c]{@{}l@{}}
\textit{Context}: my favorite city is seattle.\\ \textit{Target}: i ride my bicycle everywhere.\end{tabular}                                                                                                                                                                               \\
\hdashline
\begin{tabular}[c]{@{}l@{}}
 \Tstrut
 \textit{GPT2}: Seattle is my favorite city to go to\\ \textit{Multigen}: So what do you do when you go to the seattle\\ \textit{CODA}: I bought my bicycle from a bike shop in seattle.\\ \textit{CODA-Path}: favorite city is the location which has \\bicycle shop  is a dependency of ride bicycle\end{tabular} \\

\toprule
\begin{tabular}[c]{@{}l@{}}\textit{Context}: i am a server at a food place.\\ \textit{Target}: i eat greasy foods.\end{tabular}                                                                                                                                                                                        \\
\hdashline
\begin{tabular}[c]{@{}l@{}}
 \Tstrut
 \textit{GPT-2}: I eat healthy foods at restaurants.\\ \textit{Multigen}: I hate my food.\\ \textit{CODA}: I am a server, but I don't want to eat too much.\\ \textit{CODA-Path}: server is a person not desires eat greasy food\\
\bottomrule
\end{tabular}                                                      

\end{tabular}
 \caption{
 \small 
 Sample representative model outputs.The knowledge paths used by \coda{} provide interpretability and control over the response generation process}
    \vspace{-5mm}
    \label{tab:examples}
\end{table}

\noindent
\textbf{Path quality}: We conduct a human evaluation study to measure the quality of the generated paths. For randomly selected 100 generated responses, we ask annotators to judge 1) Relevance: Is the path relevant and used in the response? and 2) Makes sense: Does the path makes sense? Results reveal that 79\% of the paths were judged to be relevant and 76\% of the paths were judged to make sense. Thus in aggregate, the generated knowledge is good in quality, and is used in the generated response. \textbf{Path novelty}: We {analyzed the paths generated by \textsc{CODA} which were judged as sensible} by human annotators and found that
26.8\% of entities in the paths were not found in ConceptNet. This include entities such as `favorite food', `pet kitten', `single kid' and `online class'. Thus, the actual paths from the ConceptNet might not be able to cover a large fraction of head/tail entities. Furthermore,
81\% of sensible paths are novel and do not exist in ConceptNet. For example, even though the path `eat motivates go to restaurant has subevent dinner is the location for bread' exist in ConceptNet, the path `eat motivates go to restaurant has subevent dinner is the location for pizza' does not exist in ConceptNet. 
Thus we show that CODA can generalize to new entities and paths. 

In Appendix~\ref{sec:hitl} we discuss a human-in-the-loop study for controllability. The human-in-the-loop experiment shows that even minimal human intervention in the form of domain relevant keywords input for knowledge paths can improve the quality and smoothness of the transition responses. 
\section{Conclusion}
In this work, we propose and evaluate models for target-guided response generation using explicit commonsense bridging paths. 
We also introduce an automated metric to evaluate smoothness of a transition response.
We showed that our model generates more smooth and informative outputs through automatic and human evaluation. Furthermore, it allows for
more interpretable results. 
Going forward, we envision a model which could combine target and non-target guided dialogue planning. 

\section*{Acknowledgments}
We thank Maarten Sap and the anonymous reviewers for providing valuable feedback.
This work was funded by the Defense Advanced Research Planning Agency (DARPA) under DARPA
Grant N6600198-18908, and the National Science Foundation under Award No. IIS1816012. Any opinions, findings, and conclusions or recommendations expressed in this material are those of the authors and do not necessarily reflect the views of the funding agencies.

\section*{Ethics and Broader Impact}
\label{sec:ethics}

\textbf{Broader Impact and applications}:
Our proposed models for target-guided response generation can be used to generate responses based on target sentences that can drive the system's agenda in a conversation. Deploying a target-sentence guided dialogue model needs careful consideration and testing since designating a target sentence for all turns of a conversation might disrupt the natural flow of the conversation. Therefore, they can be deployed alongside existing non-target guided dialogue models that perform free-flow conversations without predesignated targets. At each turn of a conversation, a central system can use the target-coherence metric to decide if the system should generate a target-guided response or a simple follow-up response to the context.
Target-guided systems can used for several useful applications such as creating non-obtrusive recommendations, comforting people, recommending new products and services, and introducing interesting new topics and educating users about those topics.

\noindent
\textbf{Potential risks and solutions}:
We wish to raise awareness about potential misuse of proposed systems for persuading users by people with ill intentions. For example, conversational systems can pose as humans and then proactively alter user's perceptions about specific issues, evaluations of products or services, or political inclinations. To circumvent such issues, it is necessary to improve transparency through regulations, such as informing the users that they are conversing with a bot and not a human. 
Regulations are necessary to avoid hazardous outcomes during deployment for specific domains.
For example, European Union's regulatory framework proposal on artificial intelligence\footnote{\url{https://digital-strategy.ec.europa.eu/en/policies/regulatory-framework-ai}} defines use of AI systems for ``educational or vocational training, that may determine the access to education and professional course of someone’s life'' as high risk. Anyone who uses or builds upon our system should comply with such regulations. Apart from regulations, recent safety and ethics related research and datasets~\cite{baheti-etal-2021-just,sun2021safety} in conversational AI can help in mitigating aforementioned issues. \citet{10.1145/3278721.3278777} and \citet{ dinan2021anticipating} highlight and discuss potential ethical and safety issues that arise in dialogue systems research. \citet{xu2020recipes} provides a review of recent methods that try to mitigate safety issues in open-domain dialogue generation which can be utilized for the target-guided response generation task.

\noindent
\textbf{Limitations and potential biases}:
Current conversational systems suffer from several limitations, such as, they are not good at human qualities such as empathy, morality, discretion and factual correctness. There is a risk that a target driven system would ignore these factors to achieve the target. Therefore more research is needed to equip bots with such qualities. 
Our models are trained on existing datasets such as Otters and DailyDialog, and also leverage external commonsense knowledge resources. 
Knowledge graphs such as ConceptNet have been found to contain biases and have weak representations of moral common sense knowledge~\cite{hulpus-etal-2020-knowledge, mehrabi-etal-2021-lawyers}. While grounding on knowledge paths from knowledge graphs can provide insights and explanations about the model's reasoning, our models could potentially inherit biases present in these data sources. Advancements in adding a moral dimension to KGs, and extending them with intuition of morality (such as crime is bad), can enable generation of morally correct knowledge paths. Furthermore, imbuing conversational systems with empathy~\cite{MA202050}, moral discretion~\cite{ziems2022moral} and factual correctness~\cite{gupta2021dialfact,dziri2022faithdial} will improve users' experience and trust in the system.

We have included the Mechanical Turk arrangements and worker pay in the last paragraph of the section 7.3. We paid well above the US federal minimum wage (around \$13 hourly) and provided enough time to the workers to complete the task which was determined based on a few pilot experiments.

\bibliography{acl,anthology}

\begin{thebibliography}{60}
\expandafter\ifx\csname natexlab\endcsname\relax\def\natexlab#1{#1}\fi

\bibitem[{Arabshahi et~al.(2020)Arabshahi, Lee, Gawarecki, Mazaitis, Azaria,
  and Mitchell}]{arabshahi2020conversational}
Forough Arabshahi, Jennifer Lee, Mikayla Gawarecki, Kathryn Mazaitis, Amos
  Azaria, and Tom Mitchell. 2020.
\newblock Conversational neuro-symbolic commonsense reasoning.
\newblock \emph{arXiv preprint arXiv:2006.10022}.

\bibitem[{Baheti et~al.(2021)Baheti, Sap, Ritter, and
  Riedl}]{baheti-etal-2021-just}
Ashutosh Baheti, Maarten Sap, Alan Ritter, and Mark Riedl. 2021.
\newblock \href {https://aclanthology.org/2021.emnlp-main.397} {Just say no:
  Analyzing the stance of neural dialogue generation in offensive contexts}.
\newblock In \emph{Proceedings of the 2021 Conference on Empirical Methods in
  Natural Language Processing}, pages 4846--4862, Online and Punta Cana,
  Dominican Republic. Association for Computational Linguistics.

\bibitem[{Banerjee and Lavie(2005)}]{banerjee2005meteor}
Satanjeev Banerjee and Alon Lavie. 2005.
\newblock Meteor: An automatic metric for mt evaluation with improved
  correlation with human judgments.
\newblock In \emph{Proceedings of the acl workshop on intrinsic and extrinsic
  evaluation measures for machine translation and/or summarization}, pages
  65--72.

\bibitem[{Bosselut et~al.(2019)Bosselut, Rashkin, Sap, Malaviya, Celikyilmaz,
  and Choi}]{bosselut-etal-2019-comet}
Antoine Bosselut, Hannah Rashkin, Maarten Sap, Chaitanya Malaviya, Asli
  Celikyilmaz, and Yejin Choi. 2019.
\newblock \href {https://doi.org/10.18653/v1/P19-1470} {{COMET}: Commonsense
  transformers for automatic knowledge graph construction}.
\newblock In \emph{Proceedings of the 57th Annual Meeting of the Association
  for Computational Linguistics}, pages 4762--4779, Florence, Italy.
  Association for Computational Linguistics.

\bibitem[{Devlin et~al.(2019)Devlin, Chang, Lee, and
  Toutanova}]{devlin-etal-2019-bert}
Jacob Devlin, Ming-Wei Chang, Kenton Lee, and Kristina Toutanova. 2019.
\newblock \href {https://doi.org/10.18653/v1/N19-1423} {{BERT}: Pre-training of
  deep bidirectional transformers for language understanding}.
\newblock In \emph{Proceedings of the 2019 Conference of the North {A}merican
  Chapter of the Association for Computational Linguistics: Human Language
  Technologies, Volume 1 (Long and Short Papers)}, pages 4171--4186,
  Minneapolis, Minnesota. Association for Computational Linguistics.

\bibitem[{Dinan et~al.(2021)Dinan, Abercrombie, Bergman, Spruit, Hovy, Boureau,
  and Rieser}]{dinan2021anticipating}
Emily Dinan, Gavin Abercrombie, A~Stevie Bergman, Shannon Spruit, Dirk Hovy,
  Y-Lan Boureau, and Verena Rieser. 2021.
\newblock Anticipating safety issues in e2e conversational ai: Framework and
  tooling.
\newblock \emph{arXiv preprint arXiv:2107.03451}.

\bibitem[{Donahue et~al.(2020)Donahue, Lee, and
  Liang}]{DBLP:conf/acl/DonahueLL20}
Chris Donahue, Mina Lee, and Percy Liang. 2020.
\newblock \href {https://doi.org/10.18653/v1/2020.acl-main.225} {Enabling
  language models to fill in the blanks}.
\newblock In \emph{Proceedings of the 58th Annual Meeting of the Association
  for Computational Linguistics, {ACL} 2020, Online, July 5-10, 2020}, pages
  2492--2501. Association for Computational Linguistics.

\bibitem[{Dziri et~al.(2022)Dziri, Kamalloo, Milton, Zaiane, Yu, Ponti, and
  Reddy}]{dziri2022faithdial}
Nouha Dziri, Ehsan Kamalloo, Sivan Milton, Osmar Zaiane, Mo~Yu, Edoardo~M
  Ponti, and Siva Reddy. 2022.
\newblock Faithdial: A faithful benchmark for information-seeking dialogue.
\newblock \emph{arXiv preprint arXiv:2204.10757}.

\bibitem[{Ghazvininejad et~al.(2018)Ghazvininejad, Brockett, Chang, Dolan, Gao,
  Yih, and Galley}]{DBLP:conf/aaai/GhazvininejadBC18}
Marjan Ghazvininejad, Chris Brockett, Ming{-}Wei Chang, Bill Dolan, Jianfeng
  Gao, Wen{-}tau Yih, and Michel Galley. 2018.
\newblock \href
  {https://www.aaai.org/ocs/index.php/AAAI/AAAI18/paper/view/16710} {A
  knowledge-grounded neural conversation model}.
\newblock In \emph{AAAI}.

\bibitem[{Guan et~al.(2020)Guan, Huang, Zhao, Zhu, and Huang}]{Guan2020AKP}
Jian Guan, Fei Huang, Zhihao Zhao, Xiaoyan Zhu, and Minlie Huang. 2020.
\newblock A knowledge-enhanced pretraining model for commonsense story
  generation.
\newblock \emph{Transactions of the Association for Computational Linguistics},
  8:93--108.

\bibitem[{Guan et~al.(2019{\natexlab{a}})Guan, Wang, and
  Huang}]{DBLP:conf/aaai/GuanWH19}
Jian Guan, Yansen Wang, and Minlie Huang. 2019{\natexlab{a}}.
\newblock \href {https://doi.org/10.1609/aaai.v33i01.33016473} {Story ending
  generation with incremental encoding and commonsense knowledge}.
\newblock In \emph{The Thirty-Third {AAAI} Conference on Artificial
  Intelligence, {AAAI} 2019, The Thirty-First Innovative Applications of
  Artificial Intelligence Conference, {IAAI} 2019, The Ninth {AAAI} Symposium
  on Educational Advances in Artificial Intelligence, {EAAI} 2019, Honolulu,
  Hawaii, USA, January 27 - February 1, 2019}, pages 6473--6480. {AAAI} Press.

\bibitem[{Guan et~al.(2019{\natexlab{b}})Guan, Wang, and
  Huang}]{Guan_Wang_Huang_2019}
Jian Guan, Yansen Wang, and Minlie Huang. 2019{\natexlab{b}}.
\newblock \href {https://doi.org/10.1609/aaai.v33i01.33016473} {Story ending
  generation with incremental encoding and commonsense knowledge}.
\newblock \emph{Proceedings of the AAAI Conference on Artificial Intelligence},
  33(01):6473--6480.

\bibitem[{Gupta et~al.(2021{\natexlab{a}})Gupta, Tsvetkov, and
  Bigham}]{DBLP:conf/acl/GuptaTB21}
Prakhar Gupta, Yulia Tsvetkov, and Jeffrey~P. Bigham. 2021{\natexlab{a}}.
\newblock \href {https://doi.org/10.18653/v1/2021.findings-acl.338}
  {Synthesizing adversarial negative responses for robust response ranking and
  evaluation}.
\newblock In \emph{Findings of the Association for Computational Linguistics:
  {ACL/IJCNLP} 2021, Online Event, August 1-6, 2021}, volume {ACL/IJCNLP} 2021
  of \emph{Findings of {ACL}}, pages 3867--3883. Association for Computational
  Linguistics.

\bibitem[{Gupta et~al.(2021{\natexlab{b}})Gupta, Wu, Liu, and
  Xiong}]{gupta2021dialfact}
Prakhar Gupta, Chien-Sheng Wu, Wenhao Liu, and Caiming Xiong.
  2021{\natexlab{b}}.
\newblock Dialfact: A benchmark for fact-checking in dialogue.
\newblock \emph{arXiv preprint arXiv:2110.08222}.

\bibitem[{Hedayatnia et~al.(2020)Hedayatnia, Gopalakrishnan, Kim, Liu, Eric,
  and Hakkani-Tur}]{hedayatnia-etal-2020-policy}
Behnam Hedayatnia, Karthik Gopalakrishnan, Seokhwan Kim, Yang Liu, Mihail Eric,
  and Dilek Hakkani-Tur. 2020.
\newblock \href {https://aclanthology.org/2020.inlg-1.46} {Policy-driven neural
  response generation for knowledge-grounded dialog systems}.
\newblock In \emph{Proceedings of the 13th International Conference on Natural
  Language Generation}, pages 412--421, Dublin, Ireland. Association for
  Computational Linguistics.

\bibitem[{Henderson et~al.(2018)Henderson, Sinha, Angelard-Gontier, Ke, Fried,
  Lowe, and Pineau}]{10.1145/3278721.3278777}
Peter Henderson, Koustuv Sinha, Nicolas Angelard-Gontier, Nan~Rosemary Ke,
  Genevieve Fried, Ryan Lowe, and Joelle Pineau. 2018.
\newblock \href {https://doi.org/10.1145/3278721.3278777} {Ethical challenges
  in data-driven dialogue systems}.
\newblock In \emph{Proceedings of the 2018 AAAI/ACM Conference on AI, Ethics,
  and Society}, AIES '18, page 123–129, New York, NY, USA. Association for
  Computing Machinery.

\bibitem[{Hulpu{\textcommabelow{s}} et~al.(2020)Hulpu{\textcommabelow{s}},
  Kobbe, Stuckenschmidt, and Hirst}]{hulpus-etal-2020-knowledge}
Ioana Hulpu{\textcommabelow{s}}, Jonathan Kobbe, Heiner Stuckenschmidt, and
  Graeme Hirst. 2020.
\newblock \href {https://aclanthology.org/2020.starsem-1.8} {Knowledge graphs
  meet moral values}.
\newblock In \emph{Proceedings of the Ninth Joint Conference on Lexical and
  Computational Semantics}, pages 71--80, Barcelona, Spain (Online).
  Association for Computational Linguistics.

\bibitem[{Ji et~al.(2020)Ji, Ke, Huang, Wei, Zhu, and
  Huang}]{DBLP:conf/emnlp/JiKHWZH20}
Haozhe Ji, Pei Ke, Shaohan Huang, Furu Wei, Xiaoyan Zhu, and Minlie Huang.
  2020.
\newblock \href {https://doi.org/10.18653/v1/2020.emnlp-main.54} {Language
  generation with multi-hop reasoning on commonsense knowledge graph}.
\newblock In \emph{Proceedings of the 2020 Conference on Empirical Methods in
  Natural Language Processing, {EMNLP} 2020, Online, November 16-20, 2020},
  pages 725--736. Association for Computational Linguistics.

\bibitem[{Keskar et~al.(2019)Keskar, McCann, Varshney, Xiong, and
  Socher}]{DBLP:journals/corr/abs-1909-05858}
Nitish~Shirish Keskar, Bryan McCann, Lav~R. Varshney, Caiming Xiong, and
  Richard Socher. 2019.
\newblock \href {http://arxiv.org/abs/1909.05858} {{CTRL:} {A} conditional
  transformer language model for controllable generation}.
\newblock \emph{CoRR}, abs/1909.05858.

\bibitem[{Li et~al.(2017)Li, Su, Shen, Li, Cao, and
  Niu}]{li-etal-2017-dailydialog}
Yanran Li, Hui Su, Xiaoyu Shen, Wenjie Li, Ziqiang Cao, and Shuzi Niu. 2017.
\newblock \href {https://aclanthology.org/I17-1099} {{D}aily{D}ialog: A
  manually labelled multi-turn dialogue dataset}.
\newblock In \emph{Proceedings of the Eighth International Joint Conference on
  Natural Language Processing (Volume 1: Long Papers)}, pages 986--995, Taipei,
  Taiwan. Asian Federation of Natural Language Processing.

\bibitem[{Lin et~al.(2019)Lin, Chen, Chen, and Ren}]{lin-etal-2019-kagnet}
Bill~Yuchen Lin, Xinyue Chen, Jamin Chen, and Xiang Ren. 2019.
\newblock \href {https://doi.org/10.18653/v1/D19-1282} {{K}ag{N}et:
  Knowledge-aware graph networks for commonsense reasoning}.
\newblock In \emph{Proceedings of the 2019 Conference on Empirical Methods in
  Natural Language Processing and the 9th International Joint Conference on
  Natural Language Processing (EMNLP-IJCNLP)}, pages 2829--2839, Hong Kong,
  China. Association for Computational Linguistics.

\bibitem[{Lin(2004)}]{lin2004rouge}
Chin-Yew Lin. 2004.
\newblock Rouge: A package for automatic evaluation of summaries.
\newblock In \emph{Text summarization branches out}, pages 74--81.

\bibitem[{Ling et~al.(2021)Ling, Cai, Hu, Liu, Chen, and
  Chen}]{DBLP:journals/ipm/LingCHLCC21}
Yanxiang Ling, Fei Cai, Xuejun Hu, Jun Liu, Wanyu Chen, and Honghui Chen. 2021.
\newblock \href {https://doi.org/10.1016/j.ipm.2020.102392} {Context-controlled
  topic-aware neural response generation for open-domain dialog systems}.
\newblock \emph{Inf. Process. Manag.}, 58(1):102392.

\bibitem[{Liu et~al.(2016)Liu, Lowe, Serban, Noseworthy, Charlin, and
  Pineau}]{liu-etal-2016-evaluate}
Chia-Wei Liu, Ryan Lowe, Iulian Serban, Mike Noseworthy, Laurent Charlin, and
  Joelle Pineau. 2016.
\newblock \href {https://doi.org/10.18653/v1/D16-1230} {How {NOT} to evaluate
  your dialogue system: An empirical study of unsupervised evaluation metrics
  for dialogue response generation}.
\newblock In \emph{Proceedings of the 2016 Conference on Empirical Methods in
  Natural Language Processing}, pages 2122--2132, Austin, Texas. Association
  for Computational Linguistics.

\bibitem[{Lowe et~al.(2017)Lowe, Noseworthy, Serban, Angelard-Gontier, Bengio,
  and Pineau}]{lowe2017towards}
Ryan Lowe, Michael Noseworthy, Iulian~Vlad Serban, Nicolas Angelard-Gontier,
  Yoshua Bengio, and Joelle Pineau. 2017.
\newblock Towards an automatic turing test: Learning to evaluate dialogue
  responses.
\newblock In \emph{Proceedings of the 55th Annual Meeting of the Association
  for Computational Linguistics (Volume 1: Long Papers)}, pages 1116--1126.

\bibitem[{Ma et~al.(2020)Ma, Nguyen, Xing, and Cambria}]{MA202050}
Yukun Ma, Khanh~Linh Nguyen, Frank~Z. Xing, and Erik Cambria. 2020.
\newblock \href {https://doi.org/https://doi.org/10.1016/j.inffus.2020.06.011}
  {A survey on empathetic dialogue systems}.
\newblock \emph{Information Fusion}, 64:50--70.

\bibitem[{Majumder et~al.(2020)Majumder, Jhamtani, Berg{-}Kirkpatrick, and
  McAuley}]{DBLP:conf/emnlp/MajumderJBM20}
Bodhisattwa~Prasad Majumder, Harsh Jhamtani, Taylor Berg{-}Kirkpatrick, and
  Julian~J. McAuley. 2020.
\newblock \href {https://doi.org/10.18653/v1/2020.emnlp-main.739} {Like hiking?
  you probably enjoy nature: Persona-grounded dialog with commonsense
  expansions}.
\newblock In \emph{Proceedings of the 2020 Conference on Empirical Methods in
  Natural Language Processing, {EMNLP} 2020, Online, November 16-20, 2020},
  pages 9194--9206. Association for Computational Linguistics.

\bibitem[{Malaviya et~al.(2020)Malaviya, Bhagavatula, Bosselut, and
  Choi}]{malaviya2020commonsense}
Chaitanya Malaviya, Chandra Bhagavatula, Antoine Bosselut, and Yejin Choi.
  2020.
\newblock Commonsense knowledge base completion with structural and semantic
  context.
\newblock \emph{Proceedings of the 34th AAAI Conference on Artificial
  Intelligence}.

\bibitem[{Mehrabi et~al.(2021)Mehrabi, Zhou, Morstatter, Pujara, Ren, and
  Galstyan}]{mehrabi-etal-2021-lawyers}
Ninareh Mehrabi, Pei Zhou, Fred Morstatter, Jay Pujara, Xiang Ren, and Aram
  Galstyan. 2021.
\newblock \href {https://aclanthology.org/2021.emnlp-main.410} {Lawyers are
  dishonest? quantifying representational harms in commonsense knowledge
  resources}.
\newblock In \emph{Proceedings of the 2021 Conference on Empirical Methods in
  Natural Language Processing}, pages 5016--5033, Online and Punta Cana,
  Dominican Republic. Association for Computational Linguistics.

\bibitem[{Niu and Bansal(2018)}]{niu2018polite}
Tong Niu and Mohit Bansal. 2018.
\newblock Polite dialogue generation without parallel data.
\newblock \emph{Transactions of the Association for Computational Linguistics},
  6:373--389.

\bibitem[{Palmer et~al.(2005)Palmer, Gildea, and
  Kingsbury}]{palmer-etal-2005-proposition}
Martha Palmer, Daniel Gildea, and Paul Kingsbury. 2005.
\newblock \href {https://doi.org/10.1162/0891201053630264} {The {P}roposition
  {B}ank: An annotated corpus of semantic roles}.
\newblock \emph{Computational Linguistics}, 31(1):71--106.

\bibitem[{Papineni et~al.(2002)Papineni, Roukos, Ward, and
  Zhu}]{papineni2002bleu}
Kishore Papineni, Salim Roukos, Todd Ward, and Wei-Jing Zhu. 2002.
\newblock Bleu: a method for automatic evaluation of machine translation.
\newblock In \emph{Proceedings of the 40th annual meeting of the Association
  for Computational Linguistics}, pages 311--318.

\bibitem[{Qin et~al.(2020{\natexlab{a}})Qin, Ye, Tang, and
  Liang}]{DBLP:conf/aaai/QinYTL20}
Jinghui Qin, Zheng Ye, Jianheng Tang, and Xiaodan Liang. 2020{\natexlab{a}}.
\newblock \href {https://aaai.org/ojs/index.php/AAAI/article/view/6390}
  {Dynamic knowledge routing network for target-guided open-domain
  conversation}.
\newblock In \emph{The Thirty-Fourth {AAAI} Conference on Artificial
  Intelligence, {AAAI} 2020, The Thirty-Second Innovative Applications of
  Artificial Intelligence Conference, {IAAI} 2020, The Tenth {AAAI} Symposium
  on Educational Advances in Artificial Intelligence, {EAAI} 2020, New York,
  NY, USA, February 7-12, 2020}, pages 8657--8664. {AAAI} Press.

\bibitem[{Qin et~al.(2020{\natexlab{b}})Qin, Shwartz, West, Bhagavatula, Hwang,
  Bras, Bosselut, and Choi}]{DBLP:conf/emnlp/QinSWBHBBC20}
Lianhui Qin, Vered Shwartz, Peter West, Chandra Bhagavatula, Jena~D. Hwang,
  Ronan~Le Bras, Antoine Bosselut, and Yejin Choi. 2020{\natexlab{b}}.
\newblock \href {https://doi.org/10.18653/v1/2020.emnlp-main.58} {Back to the
  future: Unsupervised backprop-based decoding for counterfactual and abductive
  commonsense reasoning}.
\newblock In \emph{Proceedings of the 2020 Conference on Empirical Methods in
  Natural Language Processing, {EMNLP} 2020, Online, November 16-20, 2020},
  pages 794--805. Association for Computational Linguistics.

\bibitem[{Radford et~al.(2019)Radford, Wu, Child, Luan, Amodei, Sutskever
  et~al.}]{radford2019language}
Alec Radford, Jeffrey Wu, Rewon Child, David Luan, Dario Amodei, Ilya
  Sutskever, et~al. 2019.
\newblock Language models are unsupervised multitask learners.
\newblock \emph{OpenAI blog}, 1(8):9.

\bibitem[{Sai et~al.(2020)Sai, Mohankumar, and Khapra}]{sai2020survey}
Ananya~B Sai, Akash~Kumar Mohankumar, and Mitesh~M Khapra. 2020.
\newblock A survey of evaluation metrics used for nlg systems.
\newblock \emph{arXiv preprint arXiv:2008.12009}.

\bibitem[{Sevegnani et~al.(2021)Sevegnani, Howcroft, Konstas, and
  Rieser}]{DBLP:conf/acl/SevegnaniHKR20}
Karin Sevegnani, David~M. Howcroft, Ioannis Konstas, and Verena Rieser. 2021.
\newblock \href {https://doi.org/10.18653/v1/2021.acl-long.194} {Otters:
  One-turn topic transitions for open-domain dialogue}.
\newblock In \emph{Proceedings of the 59th Annual Meeting of the Association
  for Computational Linguistics and the 11th International Joint Conference on
  Natural Language Processing, {ACL/IJCNLP} 2021, (Volume 1: Long Papers),
  Virtual Event, August 1-6, 2021}, pages 2492--2504. Association for
  Computational Linguistics.

\bibitem[{Shi and Lin(2019)}]{shi2019simple}
Peng Shi and Jimmy Lin. 2019.
\newblock Simple bert models for relation extraction and semantic role
  labeling.
\newblock \emph{arXiv preprint arXiv:1904.05255}.

\bibitem[{Song et~al.(2019)Song, Zhang, Cui, Wang, and
  Liu}]{Song2019ExploitingPI}
Haoyu Song, W.~Zhang, Yiming Cui, Dong Wang, and T.~Liu. 2019.
\newblock Exploiting persona information for diverse generation of
  conversational responses.
\newblock In \emph{IJCAI}.

\bibitem[{Spearman(1961)}]{spearman1961proof}
Charles Spearman. 1961.
\newblock The proof and measurement of association between two things.
\newblock \emph{Appleton-Century-Crofts}.

\bibitem[{Speer et~al.(2017{\natexlab{a}})Speer, Chin, and
  Havasi}]{speersConceptnet}
Robyn Speer, Joshua Chin, and Catherine Havasi. 2017{\natexlab{a}}.
\newblock \href {https://aaai.org/ocs/index.php/AAAI/AAAI17/paper/view/14972}
  {Conceptnet 5.5: An open multilingual graph of general knowledge}.

\bibitem[{Speer et~al.(2017{\natexlab{b}})Speer, Chin, and
  Havasi}]{speer2017conceptnet}
Robyn Speer, Joshua Chin, and Catherine Havasi. 2017{\natexlab{b}}.
\newblock Conceptnet 5.5: An open multilingual graph of general knowledge.
\newblock In \emph{Proceedings of the AAAI Conference on Artificial
  Intelligence}, volume~31.

\bibitem[{Sun et~al.(2021)Sun, Xu, Deng, Cheng, Zheng, Zhou, Peng, Zhu, and
  Huang}]{sun2021safety}
Hao Sun, Guangxuan Xu, Jiawen Deng, Jiale Cheng, Chujie Zheng, Hao Zhou, Nanyun
  Peng, Xiaoyan Zhu, and Minlie Huang. 2021.
\newblock On the safety of conversational models: Taxonomy, dataset, and
  benchmark.
\newblock \emph{arXiv preprint arXiv:2110.08466}.

\bibitem[{Tang et~al.(2019)Tang, Zhao, Xiong, Liang, Xing, and
  Hu}]{DBLP:conf/acl/TangZXLXH19}
Jianheng Tang, Tiancheng Zhao, Chenyan Xiong, Xiaodan Liang, Eric~P. Xing, and
  Zhiting Hu. 2019.
\newblock \href {https://doi.org/10.18653/v1/p19-1565} {Target-guided
  open-domain conversation}.
\newblock In \emph{Proceedings of the 57th Conference of the Association for
  Computational Linguistics, {ACL} 2019, Florence, Italy, July 28- August 2,
  2019, Volume 1: Long Papers}, pages 5624--5634. Association for Computational
  Linguistics.

\bibitem[{Tao et~al.(2017)Tao, Mou, Zhao, and Yan}]{tao2017ruber}
Chongyang Tao, Lili Mou, Dongyan Zhao, and Rui Yan. 2017.
\newblock Ruber: {A}n unsupervised method for automatic evaluation of
  open-domain dialog systems.
\newblock \emph{arXiv preprint arXiv:1701.03079}.

\bibitem[{Wang et~al.(2020)Wang, Peng, Ilievski, Szekely, and
  Ren}]{wang-etal-2020-connecting}
Peifeng Wang, Nanyun Peng, Filip Ilievski, Pedro Szekely, and Xiang Ren. 2020.
\newblock \href {https://doi.org/10.18653/v1/2020.findings-emnlp.369}
  {Connecting the dots: A knowledgeable path generator for commonsense question
  answering}.
\newblock In \emph{Findings of the Association for Computational Linguistics:
  EMNLP 2020}, pages 4129--4140, Online. Association for Computational
  Linguistics.

\bibitem[{Wu et~al.(2019)Wu, Guo, Zhou, Wu, Zhang, Lian, and
  Wang}]{wu-etal-2019-proactive}
Wenquan Wu, Zhen Guo, Xiangyang Zhou, Hua Wu, Xiyuan Zhang, Rongzhong Lian, and
  Haifeng Wang. 2019.
\newblock \href {https://doi.org/10.18653/v1/P19-1369} {Proactive human-machine
  conversation with explicit conversation goal}.
\newblock In \emph{Proceedings of the 57th Annual Meeting of the Association
  for Computational Linguistics}, pages 3794--3804, Florence, Italy.
  Association for Computational Linguistics.

\bibitem[{Xu et~al.(2020)Xu, Ju, Li, Boureau, Weston, and
  Dinan}]{xu2020recipes}
Jing Xu, Da~Ju, Margaret Li, Y-Lan Boureau, Jason Weston, and Emily Dinan.
  2020.
\newblock Recipes for safety in open-domain chatbots.
\newblock \emph{arXiv preprint arXiv:2010.07079}.

\bibitem[{Yang and Klein(2021)}]{DBLP:conf/naacl/YangK21}
Kevin Yang and Dan Klein. 2021.
\newblock \href {https://doi.org/10.18653/v1/2021.naacl-main.276} {{FUDGE:}
  controlled text generation with future discriminators}.
\newblock In \emph{Proceedings of the 2021 Conference of the North American
  Chapter of the Association for Computational Linguistics: Human Language
  Technologies, {NAACL-HLT} 2021, Online, June 6-11, 2021}, pages 3511--3535.
  Association for Computational Linguistics.

\bibitem[{Yang et~al.(2019)Yang, Li, Luo, Liu, and
  Sun}]{DBLP:conf/acl/YangLLLS19}
Pengcheng Yang, Lei Li, Fuli Luo, Tianyu Liu, and Xu~Sun. 2019.
\newblock \href {https://doi.org/10.18653/v1/p19-1193} {Enhancing
  topic-to-essay generation with external commonsense knowledge}.
\newblock In \emph{Proceedings of the 57th Conference of the Association for
  Computational Linguistics, {ACL} 2019, Florence, Italy, July 28- August 2,
  2019, Volume 1: Long Papers}, pages 2002--2012. Association for Computational
  Linguistics.

\bibitem[{Zhang et~al.(2020)Zhang, Kishore, Wu, Weinberger, and
  Artzi}]{DBLP:conf/iclr/ZhangKWWA20}
Tianyi Zhang, Varsha Kishore, Felix Wu, Kilian~Q. Weinberger, and Yoav Artzi.
  2020.
\newblock \href {https://openreview.net/forum?id=SkeHuCVFDr} {Bertscore:
  Evaluating text generation with {BERT}}.
\newblock In \emph{8th International Conference on Learning Representations,
  {ICLR} 2020, Addis Ababa, Ethiopia, April 26-30, 2020}. OpenReview.net.

\bibitem[{Zhao et~al.(2017)Zhao, Zhao, and Esk{\'{e}}nazi}]{ZhaoZE17}
Tiancheng Zhao, Ran Zhao, and Maxine Esk{\'{e}}nazi. 2017.
\newblock \href {https://doi.org/10.18653/v1/P17-1061} {Learning
  discourse-level diversity for neural dialog models using conditional
  variational autoencoders}.
\newblock In \emph{Proceedings of the 55th Annual Meeting of the Association
  for Computational Linguistics, {ACL} 2017, Vancouver, Canada, July 30 -
  August 4, Volume 1: Long Papers}, pages 654--664. Association for
  Computational Linguistics.

\bibitem[{Zhong et~al.(2021)Zhong, Liu, Wang, and
  Miao}]{Zhong_Liu_Wang_Miao_2021}
Peixiang Zhong, Yong Liu, Hao Wang, and Chunyan Miao. 2021.
\newblock \href {https://ojs.aaai.org/index.php/AAAI/article/view/17712}
  {Keyword-guided neural conversational model}.
\newblock \emph{Proceedings of the AAAI Conference on Artificial Intelligence},
  35(16):14568--14576.

\bibitem[{Zhong et~al.(2019)Zhong, Wang, and Miao}]{zhong2019affect}
Peixiang Zhong, Di~Wang, and Chunyan Miao. 2019.
\newblock An affect-rich neural conversational model with biased attention and
  weighted cross-entropy loss.
\newblock In \emph{Proceedings of the AAAI Conference on Artificial
  Intelligence}, volume~33, pages 7492--7500.

\bibitem[{Zhou et~al.(2018)Zhou, Young, Huang, Zhao, Xu, and
  Zhu}]{ijcai2018-643}
Hao Zhou, Tom Young, Minlie Huang, Haizhou Zhao, Jingfang Xu, and Xiaoyan Zhu.
  2018.
\newblock \href {https://doi.org/10.24963/ijcai.2018/643} {Commonsense
  knowledge aware conversation generation with graph attention}.
\newblock In \emph{Proceedings of the Twenty-Seventh International Joint
  Conference on Artificial Intelligence, {IJCAI-18}}, pages 4623--4629.
  International Joint Conferences on Artificial Intelligence Organization.

\bibitem[{Zhou et~al.(2021{\natexlab{a}})Zhou, Gopalakrishnan, Hedayatnia, Kim,
  Pujara, Ren, Liu, and Hakkani-Tur}]{zhou-etal-2021-commonsense}
Pei Zhou, Karthik Gopalakrishnan, Behnam Hedayatnia, Seokhwan Kim, Jay Pujara,
  Xiang Ren, Yang Liu, and Dilek Hakkani-Tur. 2021{\natexlab{a}}.
\newblock \href {https://aclanthology.org/2021.sigdial-1.13}
  {Commonsense-focused dialogues for response generation: An empirical study}.
\newblock In \emph{Proceedings of the 22nd Annual Meeting of the Special
  Interest Group on Discourse and Dialogue}, pages 121--132, Singapore and
  Online. Association for Computational Linguistics.

\bibitem[{Zhou et~al.(2021{\natexlab{b}})Zhou, Gopalakrishnan, Hedayatnia, Kim,
  Pujara, Ren, Liu, and Hakkani-Tur}]{zhou2021think}
Pei Zhou, Karthik Gopalakrishnan, Behnam Hedayatnia, Seokhwan Kim, Jay Pujara,
  Xiang Ren, Yang Liu, and Dilek Hakkani-Tur. 2021{\natexlab{b}}.
\newblock Think before you speak: Using self-talk to generate implicit
  commonsense knowledge for response generation.
\newblock \emph{arXiv preprint arXiv:2110.08501}.

\bibitem[{Zhou et~al.(2021{\natexlab{c}})Zhou, Hedayatnia, Gopalakrishnan, Kim,
  Pujara, Ren, Liu, and Hakkani-Tur}]{zhou-etal-2021-think}
Pei Zhou, Behnam Hedayatnia, Karthik Gopalakrishnan, Seokhwan Kim, Jay Pujara,
  Xiang Ren, Yang Liu, and Dilek Hakkani-Tur. 2021{\natexlab{c}}.
\newblock \href {https://aclanthology.org/2021.nlp4convai-1.23} {Think before
  you speak: Learning to generate implicit knowledge for response generation by
  self-talk}.
\newblock In \emph{Proceedings of the 3rd Workshop on Natural Language
  Processing for Conversational AI}, pages 251--253, Online. Association for
  Computational Linguistics.

\bibitem[{Ziems et~al.(2022)Ziems, Yu, Wang, Halevy, and Yang}]{ziems2022moral}
Caleb Ziems, Jane~A Yu, Yi-Chia Wang, Alon Halevy, and Diyi Yang. 2022.
\newblock The moral integrity corpus: A benchmark for ethical dialogue systems.
\newblock \emph{arXiv preprint arXiv:2204.03021}.

\bibitem[{Çelikyilmaz et~al.(2020)Çelikyilmaz, Clark, and
  Gao}]{elikyilmaz2020EvaluationOT}
Asli Çelikyilmaz, Elizabeth Clark, and Jianfeng Gao. 2020.
\newblock Evaluation of text generation: A survey.
\newblock \emph{ArXiv}, abs/2006.14799.

\end{thebibliography}
\bibliographystyle{acl_natbib}

\appendix



\section{Implementation Details for CODA}
\label{sec:addtional}

\subsection{Training Details for \textsc{CODA}}
\label{sec:codadetails}
\textbf{Model training}: We code our models using Pytorch and Huggingface~\footnote{ \url{https://huggingface.co/}} library. We use validation loss to do model selection.
The KPG-wc, KPG-ht and CRG models are all based on GPT-2 small architecture. We use batch size of 10 for GPT-2 models.
We use Adam optimizer with initial learning rate of $1e-4$.
We use GeForce RTX 2080 GPUs for training models. All existing code used and datasets were CC-BY 4.0 or open sourced by original authors.

\noindent
\textbf{Decoding paths and responses}: For decoding paths using the KPG models, we use temperature of 0.7 and nucleus sampling with top-p set to 0.9. We use the same decoding strategy and hyperparameters for decoding responses using CRG model. 

\noindent
\textbf{Concept Extraction}: Entities are extracted from the context, target and response to generate and align paths using the KPG models. For a sentence s, we first extract the set of noun and verb phrases from the sentence using NLTK. We design simple grammar rules to convert some phrases to a more concise forms that are similar to the kinds of nodes present in ConceptNet,e.g., “watching the star” is converted to “watch stars”. We use NLTK's POS tagging combined with the following grammar rules: (1) Nouns and Adjectives, terminated with Nouns {<NN.*|JJ>*<NN.*>}  (2) Verb and verb phrases {<RB.?>*<VB.?>*<JJ>*<VB.?>+<VB>?}. We normalize the verbs using NLTK. The final set of entities consist of the noun and verb phrases. We exclude phrases such as ``today'', ``enough'' which are sometimes incorrectly detected as entities.

\noindent
\textbf{Sub-selecting entity pairs during training of CRG model}: For every context-target pair, we have n number of pair of head-tails entities. We score an entity pair by calculating the inverse document frequencies (computed using Gutenberg English corpus) of the entity tokens and summing up the maximum value found for a token in each entity in the pair.  For training phase, we keep the topD pairs of entities. The value of top D is selected based on validation performance and comes out typically between 1-3.

\noindent
\textbf{Knowledge graph details}: The number of nodes in the ConceptNet resource we have used\footnote{ \url{www.github.com/wangpf3/Commonsense-Path-Generator}} is 382226. We perform random walks on the graph with paths of length from 1 to 6 and get a total of 3883671 number of paths. 

\noindent
\textbf{Edges in the knowledge path}:
\label{sec:edges}
We discard some edge types which are regarded to be uninformative and offer little help for our task following~\citet{wang-etal-2020-connecting}.
They include RelatedTo, Synonym, Antonym, DerivedFrom, FormOf, EtymologicallyDerivedFrom
and EtymologicallyRelatedTo. Since the nodes in ConceptNet are directional, we also add inverse edges during path sampling. For example the path ``ecosystem <--	PartOf	<--	organism'' can be sampled as ``ecosystem \_isPartOf organism'' where the underscore indicates a reverse edge.

\noindent
\subsection{Clause Identification for Data Augmentation}
\label{sec:clause}
For \textit{target creation}, given a dialogue context $c$ and its response $r$, we first break the response $r$ into sentence clauses.
For example, given a context ``Is my booking complete?'' and the response  ``your reservation is confirmed. now i need your phone number,'', we extract a clause $t$ ``i need your phone number'' as the target candidate $t$. 
For clause extraction we use Allennlp's SRL parser~\footnote{github.com/allenai/allennlp} which is trained using a BERT-based model~\cite{shi2019simple} and is based on PropBank~\cite{palmer-etal-2005-proposition}. It identifies the arguments associated with the predicates or verbs of a sentence predicates (verbs or events) in a sentence and classifies them into roles such as agent, patient and instrument. For the example above, it identifies ``need'' as a predicate with agent ``i'' and instrument ``your number''. 

\noindent
\subsection{Data Augmentation for CODA}
We filter data from the dailydialog dataset based on a threshold set to 0.7 for data augmentation. This threshold was selected using emperical performance of thr CODA model.
For \codanocskb{} model which does not use knowledge paths, the context, target and transition response is used directly in training the CRG decoder of \codanocskb{} model. But for CODA model which uses the knowledge paths, the dailydialog data is converted to the same format as Otters data, that is, we first do entity detection on the target component of the responses as well as the the dialogue context. Then we generate a set of paths for each pair of entities. The CODA model is first trained on paths from the filtered dailydialog data and then fine-tuned on the Otters dataset which follows the same knowledge path format.
The maximum dialogue history length is set to 2 for dailydialog dataset.

\subsection{Target Coherence Metric}
\label{sec:tcagain}
In Table~\ref{tab:tcexamples}, we provide examples for \textbf{stress testing} the Target-Coherence metric. TC scores for the responses are shown in brackets.
Simply repeating or addressing either the target or context gets a low TC score. For example the response ``I like stargazing outside'' is not a smooth transition and gets a low TC score, while ``I like stargazing outside with my pet'' is a smooth transition and gets a high TC score.
In Figure~\ref{fig:tcsynth} we present an overview of the mechanisms used for generating negative samples for training the Target-Coherence metric. For negative examples, 1) Given gold response r, and context c, we sample a random negative target t', which creates a response which does not transition towards the target t, 2) Given gold response r, and target t, we sample a random negative context c', which creates a response which is not coherent to the context c, 3) Given gold context c, and target t, we either sample a random negative response r' or generate a response r' conditioned on random c' or t', which creates a response which does not transition to  target t or is coherent to context c. 




\begin{table}[t]
\centering
\small
\begin{tabular}{l}
\toprule
\begin{tabular}[c]{@{}l@{}}Context: i enjoy staring up at the sky.\\ Target: i like to spend a lot of my free time with my pet.\end{tabular}                                                                                               \\
\hdashline
\begin{tabular}[c]{@{}l@{}}
Response 1: I like stargazing outside with my pet. (0.99)\\ 
Response 2: I like stargazing outside. (0.05)\\ 
Response 3: I like walking with my pet. (0.01)\\ 
Response 4: My pet is a big star. (0.02)
\end{tabular}                                                             \\
\toprule
\begin{tabular}[c]{@{}l@{}}Context: i make blogs.\\
Target: i have a large family with babies.\end{tabular}                   \\
\hdashline
\begin{tabular}[c]{@{}l@{}}
Response 1: I want to blog about my children.(0.99)\\ 
Response 2: My family has a lot of babies. (0.05)\\ 
Response 3: My blogs are very famous. (0.01)\\ 

\end{tabular}     \\                                                       \bottomrule                                                 
\end{tabular}
 \caption{
 \small 
 Stress testing the Target-Coherence metric. We show sample responses and TC score for the responses in brackets. }
    \label{tab:tcexamples}
\end{table}

\begin{table*}[t]
\centering
\small
\begin{tabular}{l|l} 
\hline
Target & Keywords \\ 
\hline
i need your address & send money; visit; mail; send gift; send coupon \\ 
\hline
you should spend time with your friends & don't be alone; mental health; be happy; \\ 
\hline
you can try our restaurant & best ingredients ; cheapest food; free delivery \\ 
\hline
our new recipe is best selling & fat free; healthy; protein; tasty \\ 
\hline
i am the best financial advisor & get rich quickly; sound advice; money management \\ 
\hline
you should have a positive attitude & mental health; others will help; peace \\ 
\hline
we should always avoid fighting & peace; happiness; injury; understand other people \\ 
\hline
i want to come to united states & freedom ;democracy; money; job; american dream; education \\ 
\hline
everyone should get vaccinated & public health; reduce hospital burden; live longer; covid; be safe \\ 
\hline
we should donate to charity & help poor; make a difference; give assistance; feel good; social benefits \\
\hline
\end{tabular}
 \caption{
 \small 
The set of manually created targets and keyword set used for each target.}
    \vspace{-4mm}
    \label{tab:hitlinput}
\end{table*}

\section{Training Details of Baselines}
\label{sec:baselinedetails}

\textbf{Training GPT-2 Fudge model}
 \citet{DBLP:conf/naacl/YangK21} proposed a future discriminator based decoding technique. The Fudge discriminator uses a discriminator trained to distinguish good response continuations from the poor ones and guides the GPT2 based decoder towards responses that are coherent to both the source and target sentences. The Fudge discriminator needs positive and negative sample data for training. We train the discriminator to distinguish a good response from a bad (not coherent to target or context). The input to train the discriminator (a LSTM model) is the concatenation of the context sentence, followed by the target sentence and finally the tokens of a response r with tokens k. The discriminator then learns to predict 1 if the next token in the response at position k belongs to the gold response or 0 if the token is a random one. 
  We train the Fudge discriminator by preparing negative instances using the same techniques we use to train the Target-Coherence model - sampling random negative responses, responses coherent to the context but not to the target, and responses coherent to the target but not to the context. 
  
  \noindent
 \textbf{Training CS-Pretrain model}
 The model is based on the commonsense story generation model from \citet{Guan2020AKP}
 We create training data for the {CS-Pretrain} model by using the same sampled paths we use for training the KPG-wc model. The paths are converted into textual format by converting edges into text sequences. The model is only pretrained with general commonsense paths and then fine-tuned on Otters dataset in a manner similar to the GPT-2 baselines (i.e. without paths).
 Our experiments show that pretraining with commonsense model does not help with target-guided task, probably since the task needs target conditional commonsense and general commonsense knowledge only confuses the model during decoding.

\noindent
   \textbf{Training Concept-Predict}  leverages  a  concept  prediction  strategy  from \citet{DBLP:conf/aaai/QinYTL20}.  The input to the model is the context and target and it predicts a single concept based on closeness to the target. The concept is then fed as an input to the CRG model along with the context and target sentences.

\begin{figure*}[t]
    \centering
    \vspace{-.2pc}
    \includegraphics[width=0.90\textwidth]{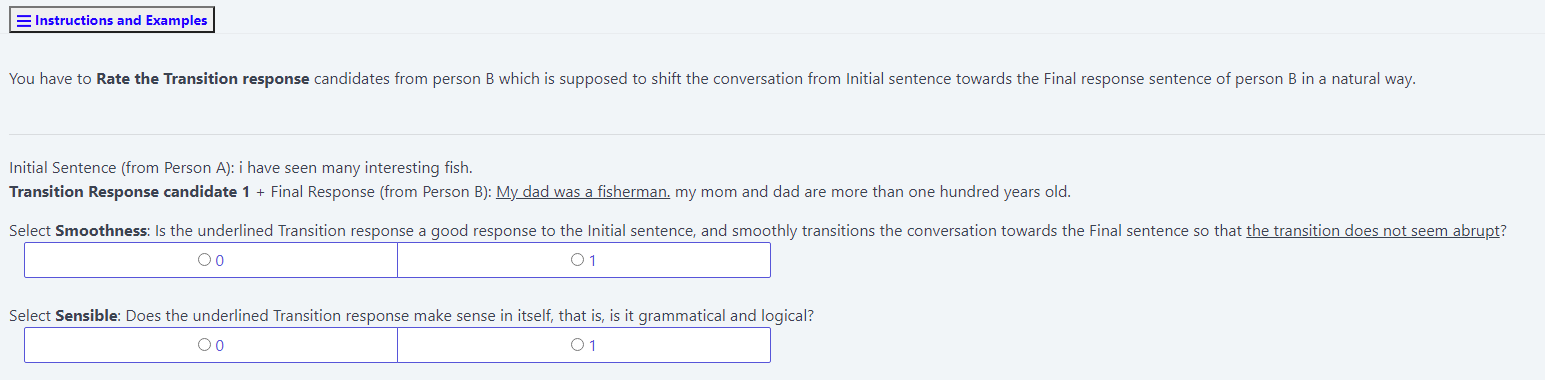}
    \vspace{-.5pc}
    \caption{
    \small 
    Amazon mechanical turk interface for human ratings collection }
    \label{fig:mturkratings}
\end{figure*}

\noindent
   \textbf{Training \codanocskb{}}: \coda{} variant that uses Dailydialog augmentation and does not use commonsense paths from KPG models in the CRG model. Therefore the model consists of only a CRG model (no KPG models) which take the context and target sentences as inputs.

\noindent
   \textbf{Training \codakw{}} \coda{} variant that uses only entities and no edges from the path. For example the path ``favorite city is the location which has bicycle shop is a dependency of ride bicycle'' is converted to ``favorite city bicycle shop ride bicycle'', which is fed as input to the CRG model.
  
\noindent
   \textbf{Training \codanoalign{}}: variant that relies on only KPG-ht for training and inference. Does not select paths based on alignment with responses. The paths used during training the CRG model come from KPG-ht instead of KPG-wc. 
   
\noindent
   \textbf{Training \codaps{}}: variant that samples paths directly from ConceptNet using the algorithm proposed in \citet{lin-etal-2019-kagnet}. Given a pair of context and target concept, we use their algorithm to sample an actual path directly from ConceptNet. The model is pretrained on Dailydialog augmented data and fine-tuned on Otters with the sampled paths from ConceptNet. The model suffers from missing entities and missing links between entities in ConceptNet which is solved by CODA.


\begin{table}[t]
\centering
\small
\begin{tabular}{l}
\toprule
\begin{tabular}[c]{@{}l@{}}Context: i dye my hair.\\ Target: we should donate to charity.\end{tabular}                                                                                                                                                                                    \\
\hdashline
\begin{tabular}[c]{@{}l@{}}
 \Tstrut
Path (KPG-oneent):  hair belongs to people motivated by\\ \underline{give assistance} has prequisite donate to charity.\\CODA-controlled: I donate my hair to a non-profit that \\ \textit{helps people in need}.\\  
 \Tstrut
Path (KPG-ht): hair belongs to people desires donate\\ to charity \\ CODA: People who donate are very good people.\\ \end{tabular}                                                             \\

\toprule
\begin{tabular}[c]{@{}l@{}}Context: i have an amazing garden.\\ Target: you can try our restaurant.\end{tabular}                                                                                                                                                                                    \\
\hdashline
\begin{tabular}[c]{@{}l@{}}
 \Tstrut
Path (KPG-oneent): garden is a location of grow food\\ motivated by goal \underline{best ingredients} is desired by person\\  capable of try restaurant \\ CODA-controlled: My restaurant uses the \textit{best ingredients}\\ from the garden.\\
 \Tstrut 
Path (KPG-ht):  garden is a location of have friends\\ over has prerequisite try restaurant\end{tabular}  \\ CODA: you can have friends over.                                                           \\

\bottomrule

\end{tabular}
 \caption{
 \small 
 Sample data and model outputs from the human-in-the-loop experiment. The underlined words are keyword inputs provided to the model KPG-oneent. The italicised words in the CODA controlled outputs are phrases are generated based on the input keywords.}
    \vspace{-5mm}
    \label{tab:hitlexamples}
\end{table}

\section{Human-in-the-loop Experiment}
\label{sec:hitl}
\textbf{Can human involvement improve generation?}
Our CRG model uses explicit paths generated from the KPG models, which not only provides interpretability, it also allows human-in-the-loop intervention for finer controllability. To test this hypothesis, we create a model KPG-oneent which is a hybrid version of KPG-wc and KPG-ht model. 
The model takes a single entity $n_k$ given by a user as an input and is trained to generate a path containing that entity.
We test this model on a manually created set of target sentences $S$ of size 10 belonging to domains such as healthcare and charity. The data created is shown in Table~\ref{tab:hitlinput}. An example sentence in set $S$ is `we should donate to charity' and we manually curate a set of keywords such as `help poor', `give assistance' and `tax deductions' that are relevant to the target sentence of interest and can guide the knowledge path sampling towards meaningful paths. This data creation took the authors 30 minutes of effort.  For 100 random sampled contexts from the Otters dataset, we select a random target sentence from the set $S$ and sample a keyword $k$ from the curated set of keywords of that target. 
We compare this controllable model with the KPG-ht model that was used for path generation in all our experiments.
We present sample outputs of the model in Table~\ref{tab:hitlexamples}. The input keywords used as intervention are underlined. The paths which use the keyword intervention generate smoother transitions compared to the paths which do not use the keyword intervention.
We find that the \textsc{Target-Coherence} metric favors the KPG-oneent model in 59 percent of cases, confirming that even minimal human intervention in the form of domain relevant keywords can improve the quality of generation.


\end{document}